\pdfoutput=1

\documentclass[11pt]{article}

\usepackage[final]{acl}

\usepackage{times}
\usepackage{latexsym}

\usepackage[T1]{fontenc}

\usepackage[utf8]{inputenc}

\usepackage{microtype}

\usepackage{inconsolata}

\usepackage{graphicx}

\usepackage{algorithm}
\usepackage[noend]{algorithmic}
\usepackage{dblfloatfix}
\usepackage{hyperref}       %
\usepackage{url}            %
\usepackage{booktabs}       %
\usepackage{amsfonts}       %
\usepackage{amssymb}
\usepackage{nicefrac}       %
\usepackage{xcolor}         %
\usepackage{array}
\usepackage{multirow}
\usepackage{multicol}
\usepackage{color, colortbl}
\usepackage{amsmath}
\usepackage{float}
\usepackage[most]{tcolorbox}
\usepackage{makecell}
\usepackage{caption}
\usepackage{xspace}
\usepackage{longtable}

\usepackage{listings}       %
\definecolor{keywordcolor}{rgb}{0.7, 0.1, 0.1}   %
\definecolor{tacticcolor}{rgb}{0.0, 0.1, 0.6}    %
\definecolor{commentcolor}{rgb}{0.4, 0.4, 0.4}   %
\definecolor{symbolcolor}{rgb}{0.0, 0.1, 0.6}    %
\definecolor{sortcolor}{rgb}{0.1, 0.5, 0.1}      %
\definecolor{attributecolor}{rgb}{0.7, 0.1, 0.1} %

\usepackage[inline]{enumitem}       %
\setlist{leftmargin=5.5mm}
\setlist[itemize]{noitemsep, topsep=0pt, parsep=0pt, partopsep=0pt}

\usepackage[colorinlistoftodos]{todonotes}
\usepackage{pgfplots}       %

\newcommand{\sparagraph}[1]{\noindent\textbf{#1.}}

\definecolor{myred}{RGB}{230, 124, 115}
\definecolor{mydarkorange}{RGB}{230, 150, 115}
\definecolor{myorange}{RGB}{245, 179, 108}
\definecolor{myyellow}{RGB}{248, 213, 103}
\definecolor{mylightgreen}{RGB}{152, 198, 124}
\definecolor{mygreen}{RGB}{111, 191, 132}
\definecolor{mylightblue}{RGB}{201, 218, 248}
\definecolor{myblue}{RGB}{140, 180, 250}
\definecolor{mydarkblue}{RGB}{110, 160, 230}
\definecolor{mygrey}{RGB}{100, 100, 100}

\newcommand{\BEqL}{BEq$_L$\xspace}
\newcommand{\BEqP}{BEq+\xspace}

\pgfplotsset{compat=1.18}

\tcbset{
    colback=mygreen!5!white,
    colframe=mygreen!90!black,
    fonttitle=\bfseries,
    breakable,
    boxrule=0.5mm,
    before upper={\parindent0pt},
    arc=0.5mm,
    enhanced jigsaw,
    top=5pt, bottom=5pt, left=5pt, right=5pt,
}

\title{Reliable Evaluation and Benchmarks for Statement Autoformalization}

\author{Auguste Poiroux \quad Gail Weiss
\\  \vspace{1mm}{\centering {\bf Viktor Kunčak \quad Antoine Bosselut}} \\
School of Computer and Communication Sciences \\
EPFL, Switzerland \\
\texttt{\{auguste.poiroux, gail.weiss, viktor.kuncak, antoine.bosselut\}@epfl.ch}}

\begin{document}
\maketitle
\begin{abstract}
    Evaluating statement autoformalization, translating natural language mathematics into formal languages like Lean 4, remains a significant challenge, with few metrics, datasets, and standards to robustly measure progress. In this work, we present a comprehensive approach combining improved metrics, robust benchmarks, and systematic evaluation, to fill this gap. First, we introduce \href{https://github.com/augustepoiroux/LeanInteract/blob/8cd9835e3aa87700886eaac2ff3e67409f544d87/examples/beq_plus.py}{\textbf{BEq+}}, an automated metric that correlates strongly with human judgment, along with \href{https://huggingface.co/datasets/PAug/ProofNetVerif}{\textbf{ProofNetVerif}}, a new dataset for assessing the quality of evaluation metrics, containing 3,752 annotated examples. Second, we develop two new autoformalization benchmarks: \href{https://huggingface.co/datasets/PAug/ProofNetSharp}{\textbf{ProofNet\#}}, a corrected version of ProofNet, and \href{https://github.com/augustepoiroux/RLMEval}{\textbf{RLM25}}, with 619 new pairs of research-level mathematics from six formalization projects. Through systematic experimentation across these benchmarks, we find that current techniques can achieve up to 45.1\% accuracy on undergraduate mathematics but struggle with research-level content without proper context. Our work establishes a reliable foundation for evaluating and advancing autoformalization systems.
\end{abstract}

\section{Introduction}

Automatic verification of logical reasoning holds promise for formal verification of mathematical proofs, software verification, and artificial intelligence. Proof assistants allow users to rigorously express mathematical statements and mechanically check their proofs \cite{assia_mahboubi_2022_7118596,DBLP:conf/mkm/Paulson23,DBLP:conf/ijcar/Avigad24}, but require inputs to be \emph{formalized}: translated from informally stated mathematical statements into a formal language, such as Lean 4. %
New research explores methods that automate this process, a task referred to as \emph{autoformalization} \citep{benzmuller_promising_2020, wang_exploration_2020}, but there exists few reliable resources for evaluating statement autoformalization methods %
\citep{liu_rethinking_2024, lu_formalalign_2024}.

To enable scalable experimentation and robust evaluation, we propose new metrics, benchmarks, and standards for evaluating statement autoformalization methods. First, we introduce \textbf{\BEqP}, a reference-based metric inspired by BEq \citep{liu_rethinking_2024}. \textbf{\BEqP} is deterministic, runs efficiently on CPU alone, and does not require a 20B LLM prover. We measure its strong correlation with human judgments, demonstrating its utility as a metric for comparing autoformalization models. Recognizing that progress in autoformalization will require continued improvements in metrics, we also release \textbf{ProofNetVerif}, a new benchmark of 3752 formal-informal pairs with human-annotated binary semantic equivalence labels that can be used to benchmark the faithfulness of new metrics.

Next, we propose two benchmarks as evaluation testbeds of statement autoformalization methods. First, we identify numerous formalization errors in existing Lean 4 ports (31.8\% of the entries), and present \textbf{ProofNet\#}, a meticulously corrected version of the leading ProofNet benchmark \citep{azerbayev_proofnet_2023}. At the same time, we observe that ProofNet focuses on undergraduate-level and self-contained statement formalization. To assess autoformalization in more realistic, context-dependent, research-oriented scenarios, we curate \textbf{RLM25}, a novel benchmark comprising 619 pairs from real-world, research-level formalization projects, the first of its kind, to our knowledge.

Finally, we set new standards for evaluating autoformalization. Current state-of-the-art methods, whether based on large language models \citep{wu_autoformalization_2022,azerbayev_proofnet_2023}, distilled back-translation \citep{jiang_multilingual_2023, azerbayev_proofnet_2023}, or retrieval augmented generation \citep{azerbayev_proofnet_2023, liu_rethinking_2024}, show limited success, with reported accuracies typically below 20\% on benchmarks such as ProofNet \citep{azerbayev_proofnet_2023, liu_rethinking_2024}. A common failure of these methods is the inability to generate formalizations that \textit{type-check} in Lean 4 \citep{10.1007/978-3-030-79876-5_37}, a crucial precursor to correctness. Our results show that sampling multiple formalizations and discarding the ones that do not type-check already substantially increases downstream correctness.

Given that statement autoformalization is more likely to be applied in limited data contexts (i.e., mathematics), increasing test-time compute is a promising factor. Consequently, we propose that autoformalization studies should explore a variety of inference-time compute budgets to broadly test method capabilities. While prior work explored sampling multiple formalizations (up to $n=20$) \citep{li_autoformalize_2024, agrawal_towards_2022}, we systematically study how performance scales with a significantly larger number of samples (up to $n = 1000$). We also investigate various strategies for filtering ill-typed samples and selecting the final prediction.

Our experiments demonstrate that scaling test-time compute by increasing candidate sampling, combined with type-check filtering and selection heuristics, can substantially improve autoformalization accuracy. Notably, the performance of GPT-4o on ProofNet\# improves from 31.0\% (single generation) to 45.1\% (using $50$ samples and a selection heuristic). These findings, alongside our contributions in metrics and benchmarks, establish a more reliable foundation for evaluating and advancing statement autoformalization systems, particularly for complex undergraduate and research-level mathematics.

\noindent
\emph{We summarize our contributions as follows:}

\begin{itemize}
    \item We develop \textbf{BEq+}, a reference-based metric for statement autoformalization based on deterministic symbolic computation and running exclusively on a CPU. We demonstrate that \textbf{BEq+ correlates strongly with human annotations at the benchmark level}, making it reliable to compare different autoformalization models.
    \item We release \textbf{ProofNetVerif}, a benchmark of $3\,752$ formal-informal pairs annotated with binary semantic equivalence labels to evaluate statement autoformalization metrics.
    \item We curate \textbf{ProofNet\#}, a revised version of ProofNet with many corrections.
    \item We release \textbf{RLM25}, a new statement autoformalization benchmark based on several research-level natural language-aligned formalization projects. To the best of our knowledge, this is the first autoformalization benchmark on research-level mathematical topics.
    \item Using \BEqP and manual evaluation, we study the statement autoformalization performance of various leading models. %
\end{itemize}

We release \href{https://github.com/augustepoiroux/LeanInteract/blob/8cd9835e3aa87700886eaac2ff3e67409f544d87/examples/beq_plus.py}{\textbf{\BEqP}}, \href{https://huggingface.co/datasets/PAug/ProofNetVerif}{\textbf{ProofNetVerif}}, and \href{https://huggingface.co/datasets/PAug/ProofNetSharp}{\textbf{ProofNet\#}} with an MIT license. \href{https://github.com/augustepoiroux/RLMEval}{\textbf{RLM25}} is released under an Apache 2.0 license, similar to the license of the underlying projects.

\section{Related Work}

\sparagraph{Interactive Theorem Proving} Autoformalization in mathematics depends on formal systems, such as Coq \citep{casteran:hal-00344237}, Lean \citep{10.1007/978-3-030-79876-5_37}, Isabelle \citep{Nipkow2002-NIPIAP}, and their math libraries. In this work, we focus on Lean (specifically, its current version, Lean 4): a powerful interactive theorem prover with a growing formal library of definitions and proven statements known as Mathlib \citep{The_mathlib_Community_2020}. %
\citet{10.1007/978-3-319-21401-6_26} and \citet{10.1007/978-3-030-79876-5_37} provide insights into the inner workings of Lean type-checking.

\sparagraph{Autoformalization} Classical programmatic tools can be used to translate \emph{constrained} natural language statements into formal systems \citep{pathak2024gflean}. In contrast, we are interested in translating \emph{unconstrained} natural language statements. In \citet{wu_autoformalization_2022}, the authors find LLMs to be a promising approach, capable of autoformalization through the use of in-context learning. In \citet{azerbayev_proofnet_2023} and \citet{jiang_multilingual_2023}, the authors demonstrate that distilled back-translation improves performance of some base models.
\citet{agrawal_towards_2022} use an advanced post-processing step to automatically fix type errors in LLM predictions. They find that, for a given problem, keeping only well-typed predictions when generating several formalization attempts is a strong filter.
In a more recent work, \citep{li_autoformalize_2024} proposes a self-consistency approach specifically designed for autoformalization. Their method clusters logically equivalent formalizations using automated theorem-proving techniques. They evaluate this approach up to $n=10$ samples per problem.

\sparagraph{Metrics} Evaluating the accuracy of models on the statement autoformalization task is a non-trivial problem. In previous works \citep{wu_autoformalization_2022, agrawal_towards_2022, azerbayev_proofnet_2023, jiang_multilingual_2023}, manual evaluation is the standard practice to report statement autoformalization performance.
Manual evaluation, though comprehensive and methodical, is a bottleneck for evaluation, motivating automatic metrics as a proxy for manual accuracy, such as BLEU \citep{wu_autoformalization_2022, azerbayev_proofnet_2023, ying_internlm-math_2024}. However, prior work \citep{azerbayev_proofnet_2023} showed that the correlation between BLEU and formalization accuracy is low. Other works have also proposed type-check rate \citep{azerbayev_proofnet_2023, agrawal_towards_2022} and symbolic equivalence metrics \citep{liu_rethinking_2024, li_autoformalize_2024} as proxies for autoformalization accuracy. In our work, we further discuss shortcomings of these metrics, and propose new metrics to overcome them.

\sparagraph{Benchmarks} ProofNet \citep{azerbayev_proofnet_2023} is a benchmark specifically designed for autoformalization. It consists of 371 undergraduate mathematical exercises, making it an essential benchmark for evaluating the performance of autoformalization models.
In a recent work on neural theorem proving \citep{hu_minictx_2024}, the authors evaluated their automated theorem prover method on research-level formal projects. Similarly, in \citet{liu_rethinking_2024}, the authors evaluate their method on Con-NF \citep{noauthor_leanprover-communitycon-nf_2025} using LLM-generated natural language statements.

\sparagraph{LLM sampling-based methods} For our evaluation, we study a method using self-consistency algorithms such as majority voting \citep{wang2023selfconsistency} and Self-BLEU \citep{zhu2018texygen}. Such methods have empirically proven to be effective across a wide range of NLP tasks \citep{li2024agents}. In particular, \citet{lewkowycz_solving_2022} demonstrated the effectiveness of the combination of sampling and majority voting on the MATH benchmark \citep{hendrycks_measuring_2021}. Further works in this direction improve over majority voting by using trained verifiers \citep{hosseini_v-star_2024}.

\section{Manual and Symbolic Metrics}

Currently, the most reliable evaluation
for autoformalization is a manual evaluation by persons with sufficient understanding of the formal proof assistant and its library. %
As in prior work, we define a formalization as \textbf{correct} if it is semantically equivalent to the provided natural language statement. Throughout this paper, \textbf{accuracy} exclusively refers to the proportion of statements evaluated as correct by \emph{manual annotation}.

\sparagraph{Symbolic Computation Metrics \BEqL and \BEqP}
\label{sec:beq_plus_description}
However, as manual evaluation is time-consuming, we propose novel automatic evaluation metrics \BEqL and \BEqP that compare a candidate formalization to a reference formalization by checking equivalence between two formulas using symbolic algorithms inside the proof assistant. We invoke proof scripts in Lean that try to prove each formula from the other. These metrics are motivated by the prior BEq metric \citep{liu_rethinking_2024} which leverages a 20B LLM trained on the theorem proving task to check equivalence.
Instead of an LLM, our metrics employ two purely symbolic, CPU-only equivalence checks that are not only more computationally efficient and interpretable, but also eliminate the need for specialized hardware, allowing us to scale our experiments to $n=1000$ samples per query.

\sparagraph{\BEqL} This metric is based on proving formula equivalence using the \lstinline{exact?} tactic (tactics can be seen as proof steps). As noted by the authors \cite{liu_rethinking_2024}, this tactic can prove equivalence for a wide variety of syntactic variations of equivalent formal statements.
\lstinline{exact?} can, in general, use other theorems from the libraries to prove the current formalization, resulting in a situation where unrelated true theorems are considered equivalent. We therefore restrict the \lstinline{exact?} tactic to only use the candidate equivalent formula. Even though this metric has a relatively high false negative rate, its simplicity makes it easier to reproduce results across works.

\sparagraph{\BEqP} To improve recall compared to \BEqL, we introduce a new symbolic metric: \BEqP. \BEqP explores several alternative proof strategies for each direction of the implication, as summarized in Algorithm~\ref{alg:beqp_algorithm}. A description of the tactics used can be found in \autoref{sec:beqp_tactics}. We apply this algorithm once for each of the two implication directions.

\begin{algorithm}[tb]
    \small
    \caption{\BEqP\ - Unidirectional}
    \label{alg:beqp_algorithm}
    \begin{algorithmic}
        \STATE \textbf{Input:} Theorem formalizations $t_1$ and $t_2$
        \STATE \textbf{Output:} Whether $t_2$ can be derived from $t_1$

        \STATE \underline{\textit{1. Run BEq\textsubscript{L}}}
        \IF{\lstinline{exact?} closes proof by using $t_1$}
        \STATE \textbf{return} TRUE
        \ENDIF

        \STATE \underline{\textit{2. Leverage conclusion matching}}
        \IF{\lstinline{apply t_1} or \lstinline{convert t_1} succeeds}
        \STATE \underline{\textit{Proving $t_1$ assumptions can be derived from $t_2$}}
        \IF{repeated applications of \lstinline{tauto}, \lstinline{simp_all_arith!}, \lstinline{noncomm_ring}, or \lstinline{exact?} close proof}
        \STATE \textbf{return} TRUE
        \ENDIF
        \ENDIF

        \STATE \underline{\textit{3. Attempt direct assumption of $t_1$}}
        \IF{\lstinline{have : goal(t_1) := by apply_rules [t_1]} succeeds}
        \IF{repeated applications of \lstinline{tauto}, \lstinline{simp_all_arith!}, \lstinline{noncomm_ring}, or \lstinline{exact?} close the subproof introduced by \lstinline{have}}
        \IF{repeated applications of \lstinline{tauto}, \lstinline{simp_all_arith!}, or \lstinline{exact? using this} close the main proof}
        \STATE \textbf{return} TRUE
        \ENDIF
        \ENDIF
        \ENDIF

        \STATE \textbf{return} FALSE
    \end{algorithmic}
\end{algorithm}

\subsection{ProofNetVerif: Benchmarking Metrics}

When proposing new automated metrics (\autoref{sec:beq_plus_description}), we must validate whether these metrics are accurate enough to approximate human evaluation. In related work \citep{liu_rethinking_2024}, the authors conduct a study on 200 sampled formalization attempts and show that BEq is accurate at the instance level, but with a relatively high amount of false negatives.

Using manual annotations from this paper, we curate a new benchmark: ProofNetVerif. Achieving high performance on this benchmark serves two key purposes: (1) establishing robust evaluation metrics for autoformalization, and (2) improving selection strategies by filtering out incorrect formalizations, thereby narrowing the search space in sampling-based methods.

ProofNetVerif contains aligned examples of a natural language statement, reference formalization, predicted formalization, and a boolean label indicating semantic equivalence between the predicted formalization and the natural language statement, making it suitable to evaluate both reference-free and reference-based metrics. The benchmark contains 3752 examples, with 1142 entries of predicted formalizations equivalent to a natural language input (the remainder being non-equivalent). With this scale, ProofNetVerif provides a challenging benchmark for designing automated metrics to evaluate autoformalization mistakes.

We report the results of our new metrics on ProofNetVerif in \autoref{tab:beq_performance_metrics_short}. In particular, we find that \BEqP largely improves over \BEqL and captures non-trivial semantical equivalence. We measured the performance of \BEqP and \BEqL only once on this benchmark and did not tune their design to fit this specific dataset. Additional details and discussions can be found in \autoref{sec:beq_instance_results}.

\begin{table}[h]
    \centering
    \caption{Binary performance, in percentage, of BEq$_L$ and BEq+ metrics on ProofNetVerif.}
    \begin{tabular}{lcc}
        \toprule
        \textbf{Metric} & \textbf{BEq$_L$} & \textbf{BEq+} \\
        \toprule
        Precision       & \textbf{100.0}   & 98.0          \\
        Recall          & 30.9             & \textbf{48.3} \\
        F1 Score        & 47.2             & \textbf{64.7} \\
        \bottomrule
    \end{tabular}
    \label{tab:beq_performance_metrics_short}
\end{table}

\subsection{Benchmark-Level Agreement}
\label{sec:metric_study}
\label{sec:metrics_correlation}

ProofNetVerif measures instance-level agreement between human evaluation and automated metrics. However, current metrics still show relatively low recall relative to near-perfect precision, underestimating true autoformalization accuracy of models.

To complement this instance-level view, we also study benchmark-level agreement, where we no longer look at whether a single formalization attempt is correct, but rather whether aggregated scores across an entire benchmark (i.e., across all problems solved by a given model under a given method) align with human judgments (i.e., accuracy). We show that \BEqP has strong correlation with accuracy: leveraging all our manual annotations on ProofNet\# conducted for this work, we measure the correlation of symbolic metrics and human evaluation at the benchmark level, amounting to 65 data points (different models and method choices, see \autoref{sec:all_results_proofnet}), and report the results in \autoref{tab:metrics_correlations}. Correlation factors between BEq+ and human evaluation are fairly high, with a Kendall coefficient of 0.872. While symbolic metrics are conservative in comparison to human-measured accuracy, this high Kendall correlation coefficient shows that \BEqP can be reliably used to compare model performance for autoformalization.

We also find that the type-check rate correlation with human evaluation is relatively low, experimentally confirming that this metric can not be used as a substitute for human evaluation. However, our empirical results in \autoref{fig:scaling_study} suggest that type-check rates can probably be used to approximate accuracy when comparing different experiments for a \textit{single} model. This can be particularly useful in reference-free setups outside of benchmarks where no reference formalizations are available.

\begin{table}
    \centering
    \setlength{\tabcolsep}{5pt}
    \caption{Correlation between human-reported accuracy and different automated metrics on ProofNet\# using data from all models evaluated in this paper.}
    \begin{tabular}{lcc}
        \toprule
        \textbf{Metric} & \textbf{Pearson} & \textbf{Kendall} \\
        \toprule
        Type-Check      & 0.655            & 0.560            \\
        BEq$_L$         & 0.966            & 0.846            \\
        BEq+            & \textbf{0.974}   & \textbf{0.872}   \\
        \bottomrule
    \end{tabular}
    \label{tab:metrics_correlations}
\end{table}

\section{New Benchmarks}
\label{sec:proofnet_1_1}

\begin{table}[tb]
    \small
    \centering
    \setlength{\tabcolsep}{1pt}
    \caption{Lean Blueprint projects used to build RLM25}
    \begin{tabular}{@{}c@{}ccr@{}}
        \toprule
        \textbf{Project}                                                                                                                     & \textbf{\#Thms} & \textbf{Lean} & \textbf{First Commit} \\ \midrule
        \href{https://github.com/fpvandoorn/carleson/tree/ec175b9008144d009269ce427b9ad43dbd70d0a5}{Carleson}                                & 111             & 4.14.0-rc2    & 20 Oct 2023           \\
        \href{https://github.com/imperialcollegelondon/FLT/tree/fed5e57b05e232f3bfe24d24098111e9dcd7bcd1}{FLT}                               & 56              & 4.14.0-rc2    & 19 Nov 2023           \\
        \href{https://github.com/pitmonticone/FLT3/tree/a199fa0467f86504a9d2f6164b0456608e586821}{FLT3}                                      & 84              & 4.7.0-rc2     & 22 Mar 2024           \\
        \href{https://github.com/teorth/pfr/tree/f6bdcac2365623d3667d3ff8fd8ddb7f95ce2313}{PFR}                                              & 145             & 4.14.0-rc3    & 13 Nov 2023           \\
        \href{https://github.com/alexkontorovich/PrimeNumberTheoremAnd/tree/6101a4b1f0cd4096b0c41cc90c7ba89f7593ef77}{PrimeNumberTheoremAnd} & 99              & 4.14.0-rc2    & 9 Jan 2024            \\
        \href{https://github.com/remydegenne/testing-lower-bounds/tree/0f09ff100a06a5e4542181514bfff74213ae126b}{testing-lower-bounds}       & 124             & 4.13.0-rc3    & 22 Feb 2024           \\ \bottomrule
    \end{tabular}
    \label{tab:real_project_details}
\end{table}

In addition to metrics, we contribute several new benchmarks for autoformalizationm, incuding RLM25, a new benchmark for research-Level mathematics, and ProofNet\#, a manual re-annotation of ProofNet that identifies and corrects a large percentage of mislabeled pairs.

\subsection{RLM25: Research-Level Mathematics}
\label{sec:rlm25_description}

To better evaluate the use of autoformalization for formalizing new mathematical results, we introduce a new benchmark, RLM25, comprised of 619 pairs natural language statement and their Lean formalizations with context\footnote{We obtained agreement from the primary authors of these projects to evaluate our models on them.} from six formalization projects (\autoref{tab:real_project_details}) that use the Lean blueprint framework \citep{massot_patrickmassotleanblueprint_2025}.
Because these projects contain natural language-aligned formalizations of research-level mathematics, they are suitable for evaluating statement autoformalization methods. We believe RLM25 is more representative of the intended use of autoformalization in mathematics research. To the best of our knowledge, we are the first to conduct such a study on real formalization projects. With 619 pairs, RLM25 is slightly larger than existing benchmarks MiniF2F (488 pairs) and ProofNet (371 pairs).

Curating this benchmark required notable engineering and analysis efforts. We use plasTeX \citep{noauthor_plastexplastex_2024} to extract natural language statements from blueprint latex files along with the Lean labels. We then use LeanDojo \cite{yang_leandojo_2023} to extract formal statements along with their context from the Lean files. Finally, we align the natural language statements with their formal counterparts using the Lean labels in the latex files. For evaluation, we use Lean REPL \citep{noauthor_leanprover-communityrepl_2025}, which we backported to previous Lean versions to make the latest features and bug fixes available for all the projects in RLM25.

The projects included in RLM25 have been selected as follows: from 11 Lean blueprint project candidates available at the time of our study, we selected 6 projects that (1) began after October 2023 to avoid data contamination, (2) contained at least 50 extractable formal-informal pairs, and (3) for which we got use agreement from their primary authors to conduct evaluations of AI systems.

\subsection{ProofNet\#: A Corrected Lean 4 ProofNet}

Past autoformalization approaches \citep{azerbayev_proofnet_2023, jiang_multilingual_2023, liu_rethinking_2024} often use the ProofNet benchmark \citep{azerbayev_proofnet_2023} for evaluation, containing 371 pairs (185 validation, 186 test) of informal statements in undergraduate mathematics and corresponding formalizations in Lean 3. As we focus on Lean 4, we start from two Lean 4 ports of ProofNet \citep{vishwakarma_rahul3613proofnet-lean4_2024, xin_deepseek-prover-v15_2024}. However, our analysis revealed that these ports, direct translations from Lean 3 to Lean 4, contained 118 entries with formalization
errors (31.8\% of the total entries), biasing downstream evaluation.
We corrected these errors, leading to a new dataset, ProofNet\# (see~\autoref{sec:proofnet_1_1_curation}), that is compatible and well-typed for Lean versions between 4.7.0 to 4.16.0-rc2. ProofNet\# remains very close to ProofNet, as only the reference formalizations are updated. Hence, the results reported in other works using reference-free metrics (e.g., human evaluation and type-check rate) remain the same.

\section{Autoformalization Methods}

In this section, we describe the leading autoformalization approaches evaluated using our new metrics and benchmarks.

\subsection{Adaptation Methods}

\sparagraph{In-context learning} On ProofNet\#, we use the 12-shot prompt from \citet{azerbayev_proofnet_2023} updated to Lean 4, which we share in \autoref{sec:few_shot_prompt}. On research-level formalization projects, similar to \citet{hu_minictx_2024} for neural theorem proving, we also consider in-file context, i.e., the content preceding the official formalizations in the project files.

\sparagraph{Fine-tuning on MMA} Empirically, it has been shown that LLMs are better at informalization, i.e., translating formal statements to informal mathematical statements, than autoformalization \citep{wu_autoformalization_2022,azerbayev_proofnet_2023}. Using this fact, \citet{jiang_multilingual_2023} informalized the Lean 4 Mathlib library with GPT-4 \citep{openai2024gpt4} to create a dataset, MMA, of formal-informal pairs that is often used for finetuning autoformalization methods.

\sparagraph{Fine-tuning on Lean Workbook} In a recent work \citep{ying_lean_workbook_2024}, the authors release a synthetically generated training set for statement autoformalization. They train a model on MiniF2F and ProofNet benchmarks, and then, through active learning, they curate a train set of $\sim 57K$ formal-informal pairs. Because of this training on the benchmarks, we only evaluate models fine-tuned on this dataset on the RLM25 benchmark.

\subsection{Sampling-Based Methods}
\label{sec:method}

We consider sampling methods designed as a plug-and-play improvement that can be seamlessly integrated with existing techniques to enhance their performance.

\sparagraph{Sampling} In our experiments, unless otherwise stated, we employ temperature sampling with $T = 0.7$ and generate $n = 50$ autoformalization attempts per informal statement.

\sparagraph{Filtering} We use Lean REPL \citep{noauthor_leanprover-communityrepl_2025} to implement our filtering step, which assesses if provided statements are well-typed. \citet{10.1007/978-3-319-21401-6_26} and \citet{10.1007/978-3-030-79876-5_37} provide detailed insights into the inner workings of the Lean type system and type-checking.

\sparagraph{Selection} In our selection process, we employ and compare four distinct heuristics to refine and choose the best outputs generated by the models: random selection, majority voting \citep{wang2023selfconsistency}, Self-BLEU \citep{zhu2018texygen}, and the symbolic equivalence method presented in \citet{li_autoformalize_2024}.
A more detailed description of each component is presented in \autoref{sec:appendix_sc_method}.

\section{Experiments}
\label{sec:experiments}

We evaluate statement autoformalization across models and methods using \BEqP and human annotations. Our experiments test (i) optimal temperature, (ii) various selection methods, (iii) how performance scales with the number of attempts, and (iv) the effect of contextual information on research-level tasks.

\subsection{Models Used in Experimental Setup}
\label{sec:models}

\sparagraph{Llemma-7B \& 34B \citep{azerbayev_llemma_2023}} These open models are based on CodeLlama 7B and 34B \citep{roziere2024code} and have been further pre-trained on the ProofPile-2 collection of mathematical data (explicitly excluding ProofNet), which was introduced along with these models. Due to their training data, these math models are particularly suited for formal-related tasks.

\sparagraph{Llama3-8B-Instruct \citep{grattafiori_llama_2024}} This is a state-of-the art open 8B model from the Llama3 family.

\sparagraph{GPT-4o \citep{openai_gpt-4o_2024}} This is a state-of-the-art general LLM. We use version \lstinline{gpt-4o-2024-05-13} for reproducibility.

\noindent To address data contamination concerns, \autoref{tab:real_project_details} includes the first commit date of each project in the RLM25 benchmark.\footnote{The authors of the projects confirmed that the first commit dates correspond to the first public appearance of these projects.} For all these projects, this commit date comes after the announced knowledge cut-off date of the models used in this work: October 2023 for GPT-4o, March 2023 for Llama3 8B, and August 2023 as the release date of Llemma 7B.

\subsection{Analysis of Sampling-Based Methods}
\label{sec:analysis_sampling}

In this section, we conduct an ablation study of various parameters involved in sampling-based methods. By default, we use a temperature of $T=0.7$, $n = 50$ samples, and 12-shot prompting in our experiments.
All the results in this section are conducted on the \textbf{validation} split of the ProofNet\# benchmark.

\paragraph{Optimal temperature}

\begin{figure}[t]
    \centering
    \definecolor{LightBlue}{RGB}{173, 216, 250}   %
\definecolor{SkyBlue}{RGB}{135, 206, 240}     %
\definecolor{MediumBlue}{RGB}{70, 130, 220}   %
\definecolor{DarkBlue}{RGB}{30, 102, 170}      %
\definecolor{DeepBlue}{RGB}{0, 51, 102}       %
\definecolor{SuperDeepBlue}{RGB}{0, 25, 51}       %

\begin{tikzpicture}
    \begin{axis}[
            name=ax1,
            width=0.28\textwidth,
            height=0.2\textheight,
            xlabel={Temperature ($T$)},
            ylabel={Score (\%)},
            xtick={0.3, 0.5, 0.7, 0.9, 1.1},
            xticklabels={0.3, 0.5, 0.7, 0.9, 1.1},
            x tick label style={font=\footnotesize},
            ylabel style={font=\small},
            ytick={0, 5, 10, 15, 20, 25},
            yticklabels={0, 5, 10, 15, 20, 25},
            y tick label style={font=\footnotesize},
            xlabel near ticks,
            ylabel near ticks,
            xmin=0.2, xmax=1.2,
            ymin=0, ymax=25,
            axis lines*=left,
            ymajorgrids=true,
            grid style=dashed,
            legend style={
                    at={(1.0,-0.35)},
                    anchor=north,
                    legend columns=3,
                    draw=none,
                    /tikz/every even column/.append style={column sep=0.3cm}
                },
            legend image post style={xscale=0.75},
        ]
        \addplot[
            draw=LightBlue,
            mark=*,
            mark size=1pt,
            thick,
            line width=1.2pt,
            mark options={fill=LightBlue, draw=LightBlue},
            smooth,
            tension=0.3
        ] coordinates {
                (0.3, 2.59) (0.5, 2.81) (0.7, 1.84) (0.9, 1.14) (1.1, 0.76)
            };

        \addplot[
            draw=SkyBlue,
            mark=*,
            mark size=1pt,
            thick,
            line width=1.2pt,
            mark options={fill=SkyBlue, draw=SkyBlue},
            smooth,
            tension=0.3
        ] coordinates {
                (0.3, 5.46) (0.5, 5.03) (0.7, 3.41) (0.9, 2.00) (1.1, 1.19)
            };

        \addplot[
            draw=MediumBlue,
            mark=*,
            mark size=1pt,
            thick,
            line width=1.2pt,
            mark options={fill=MediumBlue, draw=MediumBlue},
            smooth,
            tension=0.3
        ] coordinates {
                (0.3, 8.65) (0.5, 8.86) (0.7, 7.14) (0.9, 4.38) (1.1, 2.00)
            };

        \addplot[
            draw=DarkBlue,
            mark=*,
            mark size=1pt,
            thick,
            line width=1.2pt,
            mark options={fill=DarkBlue, draw=DarkBlue},
            smooth,
            tension=0.3
        ] coordinates {
                (0.3, 11.68) (0.5, 12.65) (0.7, 11.35) (0.9, 7.51) (1.1, 4.00)
            };

        \addplot[
            draw=DeepBlue,
            mark=*,
            mark size=1pt,
            thick,
            line width=1.2pt,
            mark options={fill=DeepBlue, draw=DeepBlue},
            smooth,
            tension=0.3
        ] coordinates {
                (0.3, 13.62) (0.5, 15.24) (0.7, 14.70) (0.9, 11.30) (1.1, 7.08)
            };

        \addplot[
            draw=SuperDeepBlue,
            mark=*,
            mark size=1pt,
            thick,
            line width=1.2pt,
            mark options={fill=SuperDeepBlue, draw=SuperDeepBlue},
            smooth,
            tension=0.3
        ] coordinates {
                (0.3, 16.76) (0.5, 18.92) (0.7, 19.46) (0.9, 17.30) (1.1, 11.35)
            };

        \legend{$n=1$, $n=2$, $n=5$, $n=10$, $n=20$, $n=50$}
    \end{axis}

    \begin{axis}[
            at={(ax1.south east)},
            name=ax2,
            xshift=0.3cm,
            xlabel={Temperature ($T$)},
            width=0.28\textwidth,
            height=0.2\textheight,
            xtick={0.3, 0.5, 0.7, 0.9, 1.1},
            xticklabels={0.3, 0.5, 0.7, 0.9, 1.1},
            x tick label style={font=\footnotesize},
            ylabel style={font=\small},
            ytick={0, 5, 10, 15, 20, 25},
            yticklabels={0, 5, 10, 15, 20, 25},
            y tick label style={font=\footnotesize},
            xlabel near ticks,
            ylabel near ticks,
            xmin=0.2, xmax=1.2,
            ymin=0, ymax=25,
            axis lines*=left,
            ymajorgrids=true,
            ymajorticks=false,
            grid style=dashed,
            xmode=linear,
        ]
        \addplot[
            draw=LightBlue,
            mark=*,
            mark size=1pt,
            thick,
            line width=1.2pt,
            mark options={fill=LightBlue, draw=LightBlue},
            smooth,
            tension=0.3
        ] coordinates {
                (0.3, 5.68) (0.5, 5.30) (0.7, 3.51) (0.9, 1.95) (1.1, 0.59)
            };

        \addplot[
            draw=SkyBlue,
            mark=*,
            mark size=1pt,
            thick,
            line width=1.2pt,
            mark options={fill=SkyBlue, draw=SkyBlue},
            smooth,
            tension=0.3
        ] coordinates {
                (0.3, 7.84) (0.5, 7.73) (0.7, 7.14) (0.9, 3.95) (1.1, 1.41)
            };

        \addplot[
            draw=MediumBlue,
            mark=*,
            mark size=1pt,
            thick,
            line width=1.2pt,
            mark options={fill=MediumBlue, draw=MediumBlue},
            smooth,
            tension=0.3
        ] coordinates {
                (0.3, 13.08) (0.5, 12.49) (0.7, 10.81) (0.9, 7.73) (1.1, 4.54)
            };

        \addplot[
            draw=DarkBlue,
            mark=*,
            mark size=1pt,
            thick,
            line width=1.2pt,
            mark options={fill=DarkBlue, draw=DarkBlue},
            smooth,
            tension=0.3
        ] coordinates {
                (0.3, 15.46) (0.5, 16.38) (0.7, 14.00) (0.9, 11.73) (1.1, 6.54)
            };

        \addplot[
            draw=DeepBlue,
            mark=*,
            mark size=1pt,
            thick,
            line width=1.2pt,
            mark options={fill=DeepBlue, draw=DeepBlue},
            smooth,
            tension=0.3
        ] coordinates {
                (0.3, 17.84) (0.5, 18.86) (0.7, 18.70) (0.9, 15.14) (1.1, 10.27)
            };

        \addplot[
            draw=SuperDeepBlue,
            mark=*,
            mark size=1pt,
            thick,
            line width=1.2pt,
            mark options={fill=SuperDeepBlue, draw=SuperDeepBlue},
            smooth,
            tension=0.3
        ] coordinates {
                (0.3, 21.08) (0.5, 23.78) (0.7, 23.78) (0.9, 21.08) (1.1, 15.68)
            };
    \end{axis}

    \node[anchor=south] at (ax1.north) {\textbf{BEq$_L$}};
    \node[anchor=south] at (ax2.north) {\textbf{BEq+}};
\end{tikzpicture}
    \caption{\textbf{Evolution of BEq+ and BEq$_L$ pass@$n$ scores for different temperatures on top of type-check filtering}. We evaluate Llemma 7B on ProofNet\# validation split.}
    \label{fig:temperature_optimal}
\end{figure}

In \autoref{fig:temperature_optimal}, we report the evolution of BEq+ pass@n metric\footnote{pass@n measures the percentage of tasks where at least one of the n outputs is correct, as measured by BEq+ here.} using the Llemma 7B model. We find that the optimal temperature for balancing exploration and coherent outputs depends on the number of samples, and that this optimal temperature increases with the number of samples. For the rest of this study, we continue with the value of $T=0.7$ for our sampling-based experiments.

\paragraph{Optimal selection method}
\label{sec:optimal_performance}

In \autoref{fig:scaling_1000}, we report BEq+ scores of the different selection methods, along with the optimal score that can be achieved with a perfect selection method, which is represented by pass@n. First, BEq+ pass@n scores steadily increase with the number of samples, going, for example, from $4.11\%$ at $n=1$, to $40.54\%$ at $n=1000$ for Llemma 7B, meaning that there is great potential in enhancing autoformalization performance through sampling. Additional studies on BEq+ pass@n can be found in \autoref{sec:beq_pass_n_rlm25}.
Second, we find that the studied self-consistency methods performance monotonically increases with the number of samples. We confirm these findings with more models and manual evaluation in \autoref{fig:scaling_study}.
Finally, we find that the symbolic equivalence method does not yield better empirical results for selection compared to simpler and less compute-intensive methods such as Self-BLEU or majority voting. We therefore continue with Self-BLEU and majority voting in subsequent experiments.

\begin{figure}[t]
    \centering
    \definecolor{mypurple}{rgb}{0.5, 0.0, 0.5}  %

\definecolor{myorange1}{rgb}{0.95, 0.85, 0.3}  %
\definecolor{myorange2}{rgb}{0.9, 0.55, 0.1}  %
\definecolor{myorange3}{rgb}{0.6, 0.1, 0.1}  %
\definecolor{myorange4}{rgb}{0.45, 0.4, 0.4}  %

\begin{tikzpicture}
    \begin{axis}[
            name=ax1,
            width=0.21\textwidth,
            height=0.19\textheight,
            xlabel={$n$},
            ylabel={BEq+ (\%)},
            xtick={1, 10, 100, 1000},
            xticklabels={1, 10, 100, 1\,000},
            x tick label style={font=\scriptsize},
            ylabel style={font=\footnotesize, at={(axis description cs:-0.16,0.5)}, anchor=south},
            ytick={0, 5, 10, 15, 20, 25, 30, 40},
            yticklabel style={font=\scriptsize},
            xlabel near ticks,
            ymin=0, ymax=27,
            axis lines*=left,
            ymajorgrids=true,
            grid style=dashed,
            xmode=log,
            legend style={
                    at={(1.7,-0.28)},
                    anchor=north,
                    legend columns=2,
                    font=\footnotesize,
                    draw=none,
                    /tikz/every even column/.append style={column sep=0.4cm}
                },
            legend image post style={xscale=0.75},
            trim axis left,
        ]
        \addplot[draw=mypurple, mark=*, mark size=1pt, thick, line width=1.2pt, mark options={fill=mypurple, draw=mypurple}, smooth, tension=0.3] coordinates {
                (1, 3.35) (2, 4.11) (5, 5.62) (10, 8.00) (20, 9.68) (50, 13.19) (100, 14.16) (200, 16.22) (500, 19.78) (1000, 23.24)
            };

        \addplot[draw=myorange1, mark=*, mark size=1pt, thick, line width=1.2pt, mark options={fill=myorange1, draw=myorange1}, smooth, tension=0.3] coordinates {
                (1, 3.35) (2, 3.24) (5, 4.38) (10, 5.41) (20, 7.30) (50, 7.78) (100, 8.27) (200, 9.03) (500, 9.46) (1000, 9.84)
            };

        \addplot[draw=myorange2, mark=*, mark size=1pt, thick, line width=1.2pt, mark options={fill=myorange2, draw=myorange2}, smooth, tension=0.3] coordinates {
                (1, 3.35) (2, 3.78) (5, 5.62) (10, 5.57) (20, 7.35) (50, 8.27) (100, 9.89) (200, 10.70) (500, 10.97) (1000,  11.68)
            };

        \addplot[draw=myorange3, mark=*, mark size=1pt, thick, line width=1.2pt, mark options={fill=myorange3, draw=myorange3}, smooth, tension=0.3] coordinates {
                (1, 3.08) (2, 3.46) (5, 5.19) (10, 5.62) (20, 6.54) (50, 8.49) (100, 9.41) (200, 10.22) (500, 10.97) (1000, 12.97)
            };

        \addplot[draw=myorange4, mark=*, mark size=1pt, thick, line width=1.2pt, mark options={fill=myorange4, draw=myorange4}, smooth, tension=0.3] coordinates {
                (1, 3.24) (2, 3.24) (5, 4.32) (10, 4.86) (20, 6.49) (50, 8.11)
            };

        \legend{Optimal (= pass@$n$), Random, Majority Voting, Self-BLEU, Symbolic Equivalence}
    \end{axis}

    \begin{axis}[
            at={(ax1.south east)},
            name=ax2,
            xshift=0.5cm,
            xlabel={$n$},
            width=0.21\textwidth,
            height=0.19\textheight,
            xtick={1, 10, 100, 1000},
            xticklabels={1, 10, 100, 1\,000},
            x tick label style={font=\scriptsize},
            ylabel style={font=\small},
            ytick={0, 10, 20, 30, 40, 50},
            yticklabel style={font=\scriptsize},
            ymin=0, ymax=45,
            axis lines*=left,
            ymajorgrids=true,
            ymajorticks=true,
            grid style=dashed,
            xmode=log,
        ]
        \addplot[draw=mypurple, mark=*, mark size=1pt, thick, line width=1.2pt, mark options={fill=mypurple, draw=mypurple}, smooth, tension=0.3] coordinates {
                (1, 4.11) (2, 6.70) (5, 11.30) (10, 15.08) (20, 19.46) (50, 24.92) (100, 28.43) (200, 32.92) (500, 37.41) (1000, 40.54)
            };

        \addplot[draw=myorange1, mark=*, mark size=1pt, thick, line width=1.2pt, mark options={fill=myorange1, draw=myorange1}, smooth, tension=0.3] coordinates {
                (1, 4.11) (2, 4.70) (5, 6.49) (10, 8.81) (20, 8.16) (50, 8.54) (100, 9.62) (200, 9.95) (500, 8.32) (1000, 9.46)
            };

        \addplot[draw=myorange2, mark=*, mark size=1pt, thick, line width=1.2pt, mark options={fill=myorange2, draw=myorange2}, smooth, tension=0.3] coordinates {
                (1, 3.57) (2, 5.46) (5, 7.73) (10, 8.49) (20, 10.11) (50, 12.16) (100, 12.86) (200, 13.84) (500, 15.03) (1000,  15.35)
            };

        \addplot[draw=myorange3, mark=*, mark size=1pt, thick, line width=1.2pt, mark options={fill=myorange3, draw=myorange3}, smooth, tension=0.3] coordinates {
                (1, 4.32) (2, 4.38) (5, 6.86) (10, 9.62) (20, 12.00) (50, 13.62) (100, 14.49) (200, 15.51) (500, 16.92) (1000, 17.30)
            };

        \addplot[draw=myorange4, mark=*, mark size=1pt, thick, line width=1.2pt, mark options={fill=myorange4, draw=myorange4}, smooth, tension=0.3] coordinates {
                (1, 3.84) (2, 5.19) (5, 8.22) (10, 10.38) (20, 11.78) (50, 13.24)
            };

    \end{axis}

    \begin{axis}[
            at={(ax2.south east)},
            name=ax3,
            xshift=0.5cm,
            xlabel={$n$},
            width=0.21\textwidth,
            height=0.19\textheight,
            xtick={1, 10, 100, 1000},
            xticklabels={1, 10, 100, 1\,000},
            x tick label style={font=\scriptsize},
            ylabel style={font=\small},
            ytick={0, 10, 20, 30, 40, 50},
            yticklabel style={font=\scriptsize},
            ymin=0, ymax=53,
            axis lines*=left,
            ymajorgrids=true,
            ymajorticks=true,
            grid style=dashed,
            xmode=log,
        ]
        \addplot[draw=mypurple, mark=*, mark size=1pt, thick, line width=1.2pt, mark options={fill=mypurple, draw=mypurple}, smooth, tension=0.3] coordinates {
                (1, 4.59) (2, 6.64) (5, 12.65) (10, 16.49) (20, 21.08) (50, 28.00) (100, 33.68) (200, 38.00) (500, 45.19) (1000, 48.65)
            };

        \addplot[draw=myorange1, mark=*, mark size=1pt, thick, line width=1.2pt, mark options={fill=myorange1, draw=myorange1}, smooth, tension=0.3] coordinates {
                (1, 4.00) (2, 5.84) (5, 7.51) (10, 8.54) (20, 9.08) (50, 8.81) (100, 9.62) (200, 10.59) (500, 10.22) (1000, 10.38)
            };

        \addplot[draw=myorange2, mark=*, mark size=1pt, thick, line width=1.2pt, mark options={fill=myorange2, draw=myorange2}, smooth, tension=0.3] coordinates {
                (1, 4.00) (2, 5.95) (5, 7.46) (10, 9.62) (20, 10.54) (50, 12.00) (100, 13.41) (200, 14.22) (500, 15.19) (1000,  15.57)
            };

        \addplot[draw=myorange3, mark=*, mark size=1pt, thick, line width=1.2pt, mark options={fill=myorange3, draw=myorange3}, smooth, tension=0.3] coordinates {
                (1, 4.40) (2, 5.65) (5, 8.29) (10, 10.18) (20, 11.92) (50, 13.65) (100, 14.55) (200, 14.99) (500, 15.84) (1000, 16.76)
            };

        \addplot[draw=myorange4, mark=*, mark size=1pt, thick, line width=1.2pt, mark options={fill=myorange4, draw=myorange4}, smooth, tension=0.3] coordinates {
                (1, 4.38) (2, 6.49) (5, 9.03) (10, 10.86) (20, 12.59) (50, 14.27)
            };

    \end{axis}

    \node[anchor=south] at (ax1.north) {\textbf{Llama3 8B}};
    \node[anchor=south] at (ax2.north) {\textbf{Llemma 7B}};
    \node[anchor=south] at (ax3.north) {\textbf{Llemma 34B}};
\end{tikzpicture}
    \caption{\textbf{Evolution of BEq+ metric for different selection methods} on top of type-check filtering. We evaluate Llama3 8B, Llemma 7B, and Llemma 34B on ProofNet\# validation split with a number of candidate samples up to $n = 1000$. Given its quadratic scaling with $n$ and high computational cost, the symbolic equivalence method is limited to $n \leq 50$ candidate samples.
    }
    \label{fig:scaling_1000}
\end{figure}

\begin{figure}[t]
    \centering
    \definecolor{mygreen1}{rgb}{0.95, 0.85, 0.0}  %
\definecolor{mygreen2}{rgb}{0.7, 0.85, 0.5}  %
\definecolor{mygreen3}{rgb}{0.35, 0.55, 0.0}  %
\definecolor{mygreen4}{rgb}{0.0, 0.1, 0.0}  %

\begin{tikzpicture}
    \begin{axis}[
            name=ax1,
            width=0.28\textwidth,
            height=0.2\textheight,
            xlabel={$n$},
            ylabel={Score (\%)},
            xtick={1, 5, 20, 50},
            xticklabels={1, 5, 20, 50},
            x tick label style={font=\scriptsize},
            ylabel style={font=\footnotesize},
            ytick={0, 20, 40, 60, 80, 100},
            yticklabel style={font=\scriptsize},
            xlabel near ticks,
            ylabel near ticks,
            ymin=0, ymax=100,
            axis lines*=left,
            ymajorgrids=true,
            grid style=dashed,
            xmode=log,
        ]
        \addplot[draw=mygreen1, mark=*, mark size=1pt, thick, line width=1.2pt, mark options={fill=mygreen1, draw=mygreen1}, smooth, tension=0.3] coordinates {
                (1, 9.8) (5, 22.4) (20, 32.8) (50, 42.1)
            };

        \addplot[draw=mygreen2, mark=*, mark size=1pt, thick, line width=1.2pt, mark options={fill=mygreen2, draw=mygreen2}, smooth, tension=0.3] coordinates {
                (1, 25.7) (5, 50.3) (20, 71.0) (50, 84.7)
            };

        \addplot[draw=mygreen3, mark=*, mark size=1pt, thick, line width=1.2pt, mark options={fill=mygreen3, draw=mygreen3}, smooth, tension=0.3, error bars/.cd, y dir=both, y explicit] coordinates {
                (1, 24.6) (5, 55.2) (20, 78.7) (50, 89.6)
            };

        \addplot[draw=mygreen4, mark=*, mark size=1pt, thick, line width=1.2pt, mark options={fill=mygreen4, draw=mygreen4}, smooth, tension=0.3] coordinates {
                (1, 33.3) (5, 47.0) (20, 55.7) (50, 65.6)
            };

    \end{axis}

    \begin{axis}[
            at={(ax1.south east)},
            name=ax2,
            xshift=0.6cm,
            xlabel={$n$},
            width=0.28\textwidth,
            height=0.2\textheight,
            xtick={1, 5, 20, 50},
            xticklabels={1, 5, 20, 50},
            x tick label style={font=\footnotesize},
            yticklabel style={font=\footnotesize},
            ytick={0, 10, 20, 30, 40, 50},
            xlabel near ticks,
            ymin=0, ymax=50,
            axis lines*=left,
            ymajorgrids=true,
            ymajorticks=true,
            grid style=dashed,
            xmode=log,
            legend style={
                    at={(-0.2,-0.28)},
                    anchor=north,
                    font=\footnotesize,
                    legend columns=2,
                    draw=none,
                    /tikz/every even column/.append style={column sep=1.0cm}
                },
            legend image post style={xscale=0.75}, %
        ]
        \addplot[draw=mygreen1, mark=*, mark size=1pt, thick, line width=1.2pt, mark options={fill=mygreen1, draw=mygreen1}, smooth, tension=0.3] coordinates {
                (1, 4.4) (5, 7.1) (20, 10.4) (50, 12.6)
            };

        \addplot[draw=mygreen2, mark=*, mark size=1pt, thick, line width=1.2pt, mark options={fill=mygreen2, draw=mygreen2}, smooth, tension=0.3] coordinates {
                (1, 6.0) (5, 10.9) (20, 19.7) (50, 23.5)
            };

        \addplot[draw=mygreen3, mark=*, mark size=1pt, thick, line width=1.2pt, mark options={fill=mygreen3, draw=mygreen3}, smooth, tension=0.3] coordinates {
                (1, 9.3) (5, 14.8) (20, 25.7) (50, 29.5)
            };

        \addplot[draw=mygreen4, mark=*, mark size=1pt, thick, line width=1.2pt, mark options={fill=mygreen4, draw=mygreen4}, smooth, tension=0.3] coordinates {
                (1, 25.7) (5, 34.4) (20, 38.8) (50, 44.8)
            };

        \legend{Llama3 8B, Llemma 7B, Llemma 34B, GPT-4o}
    \end{axis}

    \node[anchor=south] at (ax1.north) {\textbf{Type-Check}};
    \node[anchor=south] at (ax2.north) {\textbf{Accuracy}};
\end{tikzpicture}
    \vspace{-1em}
    \caption{\textbf{Type-Check rate and Accuracy scaling trends with respect to the number of samples} on ProofNet\# validation split using 12-shot prompting and a sampling-based method (type-check filter + Self-BLEU). The number of samples varies from \textit{n} = 1 to 50 (where Llemma 34B has top type-check rate and GPT-4o top accuracy).
        Numbers are in \autoref{tab:results_sampling_scaling}.}
    \label{fig:scaling_study}
\end{figure}

\paragraph{Type-Check Filtering}
\label{sec:ablation_study}

\begin{figure}[t]
    \centering
    \definecolor{myteal1}{rgb}{0.2, 0.8, 0.8}  %
\definecolor{myteal2}{rgb}{0.0, 0.6, 0.6}  %
\definecolor{myteal3}{rgb}{0.0, 0.4, 0.4}  %
\definecolor{mybluebis}{rgb}{0.4, 0.5, 0.8}
\definecolor{myorange1}{rgb}{1.0, 0.6, 0.2}
\definecolor{myorange4}{rgb}{0.6, 0.3, 0.0}  %

\begin{tikzpicture}

    \begin{axis}[
            name=ax2,
            xshift=0.7cm,
            ybar=0pt,
            bar width=6pt,
            width=0.5\textwidth,
            height=0.17\textheight,
            enlarge x limits={abs=5*\pgfplotbarwidth},
            symbolic x coords={Llama3 8B, Llemma 7B, Llemma 34B, GPT-4o},
            xtick=data,
            x tick label style={font=\scriptsize, rotate=45, anchor=east},
            yticklabel style={font=\scriptsize},
            ytick={0, 10, 20, 30, 40, 50},
            ylabel={Accuracy (\%)},
            ymin=0, ymax=50,
            axis lines*=left,
            ymajorgrids=true,
            ymajorticks=true,
            grid style=dashed,
            area legend,
            legend style={
                    at={(0.45,-0.55)},
                    anchor=north,
                    legend columns=2,
                    draw=none,
                    font=\scriptsize,
                    row sep=0.05cm,
                    /tikz/every even column/.append style={column sep=0.2cm}
                },
            legend cell align={left},
            legend image post style={xscale=0.75}, %
        ]
        \addplot[fill=mybluebis, draw opacity=0] coordinates {(Llama3 8B, 5.5) (Llemma 7B, 8.7) (Llemma 34B, 16.9) (GPT-4o, 26.2)};
        \addplot[fill=myteal1, draw opacity=0] coordinates {(Llama3 8B, 4.4) (Llemma 7B, 6.0) (Llemma 34B, 9.3) (GPT-4o, 25.7)};
        \addplot[fill=myteal2, draw opacity=0] coordinates {(Llama3 8B, 5.5) (Llemma 7B, 10.9) (Llemma 34B, 10.9) (GPT-4o, 29.5)};
        \addplot[fill=myteal3, draw opacity=0] coordinates {(Llama3 8B, 6.0) (Llemma 7B, 14.2) (Llemma 34B, 14.2) (GPT-4o, 28.4)};
        \addplot[fill=myorange1, draw opacity=0] coordinates {(Llama3 8B, 9.8) (Llemma 7B, 16.9) (Llemma 34B, 21.3) (GPT-4o, 43.2)};
        \addplot[fill=myorange4, draw opacity=0] coordinates {(Llama3 8B, 12.6) (Llemma 7B, 23.5) (Llemma 34B, 29.5) (GPT-4o, 44.8)};

        \legend{Greedy baseline, No filter + Random selection, No filter + Majority voting, No filter + Self-BLEU, Filter + Random selection, Filter + Self-BLEU}
    \end{axis}

    \node[anchor=south] at (ax2.north) {};
\end{tikzpicture}
    \caption{\textbf{Type-Check Filtering Ablation Study}. Accuracy scores are reported on ProofNet\# validation split for various ablations of sampling-based methods.
        More details and exact numbers are reported in \autoref{tab:results_ablation_study}.}
    \label{fig:ablation_study}
\end{figure}

In \autoref{fig:ablation_study}, we empirically study the contribution of the filtering component in sampling-based methods by evaluating with and without filtering, as well as with different selection heuristics.
While majority voting (\textit{No filter + Majority voting}) and Self-BLEU selection (\textit{No filter + Self-BLEU}) generally improve the accuracy
of random sampling, both struggle to increase the performance of random sampling beyond that of the greedy decoding baseline.
Meanwhile, adding type-check filtering substantially outperforms the greedy decoding baseline even without any final selection heuristic (\textit{Filter + Random selection}).
We conclude that the type-check filter is a critical component in sampling-based methods and that it should be applied before selection.

\subsection{Empirical Analysis on RLM25}
\label{sec:real_world_results}

\begin{table}[h]
    \centering\small
    \setlength{\tabcolsep}{3pt}
    \caption{Results, in percentage, on RLM25 for different methods and models using only the natural language statement as input (i.e., no in-file context is provided).}
    \begin{tabular}{@{}ll@{}ccc@{}}
        \toprule
        \textbf{Model} & \textbf{Method} & \textbf{Type-Check} & \textbf{BEq$_L$} & \textbf{BEq+}    \\ \toprule
        \multirow{2}{*}{Llama3 8B}
                       & 12-shot         & 2.80                & \underline{0.20} & 0.40             \\
                       & MMA             & 3.18                & 0.00             & \underline{0.54} \\ \midrule
        \multirow{3}{*}{Llemma 7B}
                       & 12-shot         & 5.06                & \textbf{2.78}    & \textbf{2.78}    \\
                       & MMA             & 8.65                & 0.79             & 1.16             \\
                       & Lean Workbook   & 23.06               & 0.17             & 1.13             \\ \midrule
        Llemma 34B
                       & 12-shot         & 4.94                & 0.79             & 0.79             \\ \midrule
        GPT-4o
                       & 12-shot         & 10.55               & 0.57             & 1.50             \\ \bottomrule
    \end{tabular}
    \label{tab:real_world_no_context_results}
\end{table}

To further confirm our findings and evaluate current autoformalization methods in more realistic settings, we conduct studies on RLM25 in this section.
We start by conducting an initial study using the current approach in the literature for statement autoformalization: only the natural language statement is provided as input to the methods. We report our results in \autoref{tab:real_world_no_context_results}. These results are very similar to the ones presented in \citet{liu_rethinking_2024} on their semi-synthetic Con-NF benchmark, with BEq results close to 0\% for all methods. %

A manual analysis of the predictions quickly reveals a key issue: without in-file context, autoformalization methods lack access to crucial information. While some missing information from a Lean file, such as local definitions, can be tackled through current retrieval-based methods, others, such as opened namespaces and local variables are still missing. To assess the importance of different file components, we conducted an ablation study to identify which contextual elements help LLMs generate accurate formalizations. As shown in \autoref{tab:file_context}, the best performance across models is achieved when proofs are removed while retaining all other contextual elements. In contrast, removing both proofs and theorems significantly degrades performance, particularly for less capable models.

\begin{table}[h]
    \centering
    \fontsize{7.5}{9.5}\selectfont
    \setlength{\tabcolsep}{2pt}
    \caption{\textbf{Ablation study on prompt content} using various models on RLM25. We report BEq+ (\%) performance. In the prompt column, '-' represents removal from the context.}
    \begin{tabular}{@{}lcccc@{}}
        \toprule
        \textbf{Prompt}      & \textbf{Llama3 8B} & \textbf{Llemma 7B} & \textbf{Llemma 34B} & \textbf{GPT-4o} \\ \toprule
        12-shot              & 0.40               & 2.78               & 0.79                & 1.50            \\
        Full file context    & 18.67              & 22.15              & 25.99               & 20.64           \\
        - theorems \& proofs & 6.60               & 13.82              & 15.63               & 17.20           \\
        - proofs             & \textbf{20.29}     & \textbf{24.16}     & \textbf{30.33}      & \textbf{24.56}  \\
        \bottomrule
    \end{tabular}
    \label{tab:file_context}
\end{table}

In \autoref{tab:file_context}, we find that in-file context prompting substantially improves performance across all methods and models compared to \autoref{tab:real_world_no_context_results}. Furthermore, fine-tuning on existing autoformalization datasets does not yield improvement over base models on RLM25, suggesting that context-aware formalization datasets are needed for tackling autoformalization for research-level projects.

\subsection{Final Results}
\label{sec:results}

We conducted a detailed baseline study along with human evaluation of the different autoformalization methods on ProofNet\#, which we report in the appendix in \autoref{tab:results_baseline_proofnet}.
Confirming results from prior works \citep{jiang_multilingual_2023, azerbayev_proofnet_2023}, we note the large proportion of errors for all methods due to type-check failures.
In \autoref{tab:full_results}, we report side-by-side final performance results between classical greedy decoding and a sampling-based method using Self-BLEU on both ProofNet\# and RLM25. These results confirm the consistent performance improvement of sampling for all tested models on the two benchmarks, ProofNet\# and RLM25.
We also report the effect of sampling on MMA and Lean Workbook fine-tuned models in \autoref{tab:results_filter_select_proofnet}. However, we find that applying the sampling strategy on base models with few-shot learning achieves better absolute accuracy.
We report additional results on other benchmarks such as PDA \citep{lu_process-driven_2024} and MiniF2F \citep{zheng_minif2f_2022} in \autoref{sec:minif2f_pda_results}.

\begin{table}[h]
    \centering\small
    \setlength{\tabcolsep}{1.5pt}
    \caption{\textbf{Performance comparison between greedy decoding and a sampling-based method on ProofNet\# and the new RLM25 benchmark}. 12-shot is used for ProofNet\#, and in-file context with proofs removed is used for RLM25.
        Sampled with $T=0.7$ and $n=50$.}
    \begin{tabular}{@{}ll@{}cc|c@{}}
        \toprule
        \multirow{2}{*}{\textbf{Model}} & \multirow{2}{*}{\textbf{Method}} & \multicolumn{2}{c}{\textbf{ProofNet\#}} & \textbf{RLM25}                      \\
        \cmidrule{3-5}
                                        &                                  & \textbf{Accuracy}                       & \textbf{BEq+}    & \textbf{BEq+}    \\
        \toprule
        \multirow{2}{*}{Llama3 8B}
                                        & Greedy                           & 3.3                                     & 3.3              & 20.3             \\
                                        & Filter + Self-BLEU               & \underline{12.0}                        & \underline{9.2}  & \underline{23.9} \\ \midrule
        \multirow{2}{*}{Llemma 7B}
                                        & Greedy                           & 10.9                                    & 6.5              & 24.2             \\
                                        & Filter + Self-BLEU               & \underline{29.3}                        & \underline{17.9} & \underline{28.8} \\ \midrule
        \multirow{2}{*}{Llemma 34B}
                                        & Greedy                           & 12.5                                    & 7.1              & 30.3             \\
                                        & Filter + Self-BLEU               & \underline{28.3}                        & \underline{14.7} & \textbf{33.4}    \\ \midrule
        \multirow{2}{*}{GPT-4o}
                                        & Greedy                           & 31.0                                    & 18.5             & 24.6             \\
                                        & Filter + Self-BLEU               & \textbf{45.1}                           & \textbf{23.4}    & \underline{31.6} \\ \bottomrule
    \end{tabular}
    \label{tab:full_results}
\end{table}

\section{Conclusion}

Our work advances and standardizes the evaluation of statement autoformalization. We introduced several key resources: \textbf{\BEqP}, an automated equivalence-checking metric demonstrating a strong correlation with human judgments, and \textbf{ProofNetVerif}, a dataset of 3752 annotated examples to validate new metrics. Alongside these, we published \textbf{ProofNet\#}, a corrected version of the popular ProofNet benchmark, and \textbf{RLM25}, the first benchmark for research-level mathematics formalization across six projects. Our experiments using these new resources reveal critical insights. Sampling strategies, combined with incorporating type-checking and selection heuristics, substantially boost performance, achieving up to 45.1\% accuracy on undergraduate mathematics (ProofNet\#), suggesting autoformalization evaluations should integrate test-time compute budgets and simple heuristic correctors for recording more realistic performance measures. %
These benchmarks and our metric provide the community with more reliable tools to measure progress. We believe these contributions will foster the development of more capable and practical autoformalization systems.

\section*{Limitations}
\label{sec:limitations}

We identify key limitations of our work. First, our newly introduced RLM25 benchmark is focused on statement autoformalization, while the underlying projects from which we construct the benchmark would support extension to proof autoformalization, which is not tackled in our work. Second, we observe that \BEqP and \BEqL face challenges in proving equivalence for long formal statements (as shown in \autoref{sec:beq_instance_results}). The recall performance of \BEqP drops significantly from 62.5\% on short statements to 29.6\% on long statements, indicating that it over-penalizes longer formal expressions. These metrics also require accurate ground truth formalizations, which are not available when formalizing completely new mathematics.

\sparagraph{Data contamination} Our in-depth study in \autoref{sec:appendix_contamination} suggests that data contamination is unlikely among the models we evaluated.

\section*{Acknowledgements} %
We thank the Lean community for their support and feedback, in particular the authors of the Lean blueprint projects included in RLM25. We thank Simon Sorg for finding and sharing an exploit of the \BEqP metric we address in \autoref{sec:beq_failure_cases}.
We also gratefully acknowledge the support of the IC school of computer and communication sciences, the Swiss National Science Foundation (No. 215390), Innosuisse (PFFS-21-29), the EPFL Center for Imaging, Sony Group Corporation, and a Meta LLM Evaluation Research Grant.

\bibliography{custom}

\appendix
\section{Appendix}

\subsection{Tactics used in \BEqP}
\label{sec:beqp_tactics}

To develop \BEqP, we checked existing tactics in Lean and Mathlib using the list provided at \href{https://github.com/haruhisa-enomoto/mathlib4-all-tactics/blob/main/all-tactics.md}{https://github.com/haruhisa-enomoto/mathlib4-all-tactics/blob/main/all-tactics.md}. We provide a brief description of the tactics we ended up using in \BEqP using the official Mathlib documentation \citep{The_mathlib_Community_2020}:

\begin{itemize}
    \item \lstinline{exact?}: Searches environment for definitions or theorems that can solve the goal using exact with conditions resolved by \lstinline{solve_by_elim}. While we are not directly interested in searching the library, this tactic is also capable of handling a few transformations on the conclusion, but also variable/hypothesis assignment. For both \BEqL and \BEqP, if \lstinline{exact?} succeeds we check that it is using the other formalization to close the goal. Otherwise it could lead to false positives.

    \item \lstinline{apply e} tries to match the current goal against the conclusion of \lstinline{e}'s type. If it succeeds, then the tactic returns as many subgoals as the number of premises that have not been fixed by type inference or type class resolution.

    \item \lstinline{convert}: The \lstinline{exact e} and \lstinline{refine e} tactics require a term \lstinline{e} whose type is definitionally equal to the goal. \lstinline{convert e} is similar to \lstinline{refine e}, but the type of \lstinline{e} is not required to exactly match the goal. Instead, new goals are created for differences between the type of \lstinline{e} and the goal using the same strategies as the \lstinline{congr!} tactic. We use this tactic to try partial matching with the conclusion of the other theorem. In particular we use the \lstinline{convert using n} variation, where \lstinline{n} determines the matching depth. We vary \lstinline{n} between 0 and 5.

    \item \lstinline{tauto} breaks down assumptions of the form \lstinline{_ ∧ _}, \lstinline{_ ∨ _}, \lstinline{_ ↔ _} and \lstinline{∃ _, _} and splits a goal of the form \lstinline{_ ∧ _}, \lstinline{_ ↔ _} or \lstinline{∃ _, _} until it can be discharged using \lstinline{reflexivity} or \lstinline{solve_by_elim}.

    \item \lstinline{simp_all_arith!}: simplifies multiple times target and all (propositional) hypotheses using the other hypotheses. Additionally, it uses normalization by linear arithmetic.

    \item \lstinline{noncomm_ring}: A tactic for simplifying identities in not-necessarily-commutative rings. It is pretty general and works on all types having a ring structure.

    \item \lstinline{have : t := ...} adds the hypothesis \lstinline{this : t} to the current goal.

    \item \lstinline{apply_rules [l₁, l₂, ...]} tries to solve the main goal by iteratively applying the list of lemmas \lstinline{[l₁, l₂, ...]} or by applying a local hypothesis. If apply generates new goals, \lstinline{apply_rules} iteratively tries to solve those goals. \lstinline{apply_rules} will also use \lstinline{rfl}, \lstinline{trivial}, \lstinline{congrFun} and \lstinline{congrArg}.
\end{itemize}

\subsection{Annotation process}
\label{sec:annotation_process}
\label{sec:proofnet_1_1_curation}

For all the evaluations presented in this paper, along with the curation of ProofNetVerif and ProofNet\#, we adopted the following systematic annotation process for each formalized statement to determine its validity:

\begin{itemize}
    \item Ill-typed statements and statements with counter-examples found by \lstinline{plausible} / \lstinline{slim_check} were immediately marked as invalid.
    \item \textbf{Manual pass through all the statements.} Hypotheses and conclusions were carefully inspected for completeness and semantic correctness, correct handling of default Lean types, correct use of definitions/instances, ... A non-exhaustive list of issues that were looked for includes:
          \begin{itemize}
              \item Missing/extra invalid hypotheses
              \item Invalid implicit types (ex: $3 / 2 = 1$ in Lean because 3 and 2 are natural numbers by default)
              \item Natural numbers (they start at 0 in Lean)
              \item Parentheses, especially in quantified propositional formulas
              \item Instances/definitions used: they must have the same semantic meaning as the ones in the informal statement to formalize
              \item Semantically invalid hypotheses
              \item Stronger / more restrictive hypotheses
              \item In case of doubts, the prediction was marked as invalid
          \end{itemize}
    \item (ProofNet\# only) Running DeepSeek-Prover-V1.5 \citep{xin_deepseek-prover-v15_2024} to find proofs on the current iteration of ProofNet. We then manually analyze these proofs to check if flaws in the formalizations not detected by the manual pass have been exploited. This step helped finding a total of 2 additional formalization mistakes.
\end{itemize}

\vspace{1em}

To reduce annotation mistakes and biases:
\begin{itemize}
    \item All predictions were anonymized with respect to the model and method that produced them. This way, the annotator was not biased in favor of some models.
    \item All predictions related to a specific input problem were evaluated together, meaning that the annotator was annotating a few dozen predictions at the same time, giving them a good overview and leading to a more objective evaluation.
    \item Each sample has been annotated twice for model evaluations and the curation of ProofNetVerif, and three times for ProofNet\#.
    \item ProofNet\# and ProofNetVerif have been publicly shared with the Lean community several months before the reviewing process, including a website to explore ProofNet\#. No errors have been reported during this period.
\end{itemize}

This annotation process was conducted by a single annotator and required a total of $\sim 200$ human-hours for the results and datasets presented in this paper.

When curating ProofNet\#, this annotation process led to the discovery of mistakes in 118 entries out of the 371 of the initial ProofNet benchmark.

\subsection{Sampling-Based Method}
\label{sec:appendix_sc_method}

The method is composed of three steps: (1) sampling, (2) type-check filtering, and then (3) selecting. We present them in this section.

\subsubsection{Sampling}

In our experiments, unless otherwise stated, we employ temperature sampling with $T = 0.7$ and generate $n = 50$ autoformalization attempts per informal statement. Depending on the models, we use the vLLM library \citep{kwon2023efficient} or the OpenAI API to generate predictions.

\textbf{Cleaning:} Certain models often try to provide proofs after generating formal statements. Furthermore, we find that generated names for theorems sometimes clash with names in the Mathlib library. To avoid being considered invalid by the Lean type-checker, we trim proofs, substitute theorem names for dummy identifiers, and normalize whitespace when parsing the generated theorems. Additionally, the Lean proof assistant requires theorems to be accompanied by proofs. To address this, we append a safe dummy \lstinline{sorry} proof to each theorem (which indicates to Lean that the proof will be provided later).

\subsubsection{Filtering}

We use Lean REPL \citep{noauthor_leanprover-communityrepl_2025} to implement our filtering step. For any formal statement, if the statement is valid, REPL will return \lstinline{declaration uses `sorry'}, which means that the statement is well-typed and that we should provide an actual proof instead of \lstinline{sorry}. Otherwise, the tool will return error messages explaining why the formal statement is ill-formed, which we use as an indicator to filter out such statements. \citet{10.1007/978-3-319-21401-6_26} and \citet{10.1007/978-3-030-79876-5_37} provide detailed insights into the inner workings of the Lean type system and type-checking.

\subsubsection{Selection}
\label{sec:selection_step}

In our selection process, we employ and compare four distinct heuristics to refine and choose the best outputs generated by the models: random selection, majority voting \citep{wang2023selfconsistency}, Self-BLEU \citep{zhu2018texygen}, and the symbolic equivalence method presented in \citet{li_autoformalize_2024}.

\sparagraph{Random:} As a baseline strategy, we randomly choose an output from the set of generated candidates.

\sparagraph{Majority voting} \citep{wang2023selfconsistency}: We aggregate multiple outputs and select the most frequently occurring candidate as the final choice, relying on consensus to mitigate the impact of any single erroneous output. Our cleaning process after the sampling step normalize the generated outputs, increasing the chance of exact string match between the predictions.

\sparagraph{Self-BLEU} \citep{zhu2018texygen}: We evaluate the similarity of the generated outputs by calculating the BLEU score between all pairs of candidates. We then select the generated candidate with the highest aggregated BLEU score.

\sparagraph{Symbolic Equivalence} \citep{li_autoformalize_2024}: the core idea of this method is to compute equivalence classes of the generated predictions, using logical equivalence. A prediction from the largest equivalence class is then selected as final prediction. As the original work has been conducted in the Isabelle formal language \citep{Nipkow2002-NIPIAP}, no implementation of this method is available in Lean. We therefore implemented a Lean version relying on our BEq+ method to compute the equivalence between statements.

\subsection{Compute usage \& cost}
\label{sec:compute_usage}

All experiments were run on 1xH100 for 7-8B models and 2xH100 for Llemma-34B. The most compute-intensive experiments, i.e., generating 1000 candidates per problem in ProofNet\#, required less than 10 hours (wall-clock time). For cost reasons, we did not sample n = 1000 candidates with GPT-4o, and the largest GPT-4o results in the paper use n = 50, costing $\sim$ \$25 per evaluation on ProofNet\#.

\subsection{BEq$_L$ \& BEq+ Performance on ProofNetVerif}
\label{sec:beq_instance_results}

\begin{table*}[ht]
    \centering
    \caption{Binary performance, in percentage, of BEq$_L$ and BEq+ metrics on ProofNetVerif. The evaluation is performed on 3 splits based on the reference formalizations length. These splits contain roughly 1250 entries each.}
    \begin{tabular}{llcc}
        \toprule
        \textbf{Reference formalization length} & \textbf{Binary Metric} & \textbf{BEq$_L$} & \textbf{BEq+} \\
        \midrule
        \multirow{3}{*}{Less than 115 characters}
                                                & Precision              & \textbf{100.0}   & 97.1          \\
                                                & Recall                 & 39.9             & \textbf{62.5} \\
                                                & F1 Score               & 57.0             & \textbf{76.1} \\
        \arrayrulecolor{lightgray}\midrule
        \multirow{3}{*}{Between 115 and 165 characters}
                                                & Precision              & \textbf{100.0}   & 99.2          \\
                                                & Recall                 & 28.2             & \textbf{41.0} \\
                                                & F1 Score               & 44.0             & \textbf{58.0} \\
        \arrayrulecolor{lightgray}\midrule
        \multirow{3}{*}{More than 165 characters}
                                                & Precision              & 100.0            & 100.0         \\
                                                & Recall                 & 17.2             & \textbf{29.6} \\
                                                & F1 Score               & 29.3             & \textbf{45.6} \\
        \arrayrulecolor{black}\midrule
        \multirow{3}{*}{All}
                                                & Precision              & \textbf{100.0}   & 98.0          \\
                                                & Recall                 & 30.9             & \textbf{48.3} \\
                                                & F1 Score               & 47.2             & \textbf{64.7} \\
        \bottomrule
    \end{tabular}
    \label{tab:beq_performance_metrics}
\end{table*}

In \autoref{tab:beq_performance_metrics}, we report the performance of our metrics at the instance level using binary metrics. While BEq+ outperforms BEq$_L$, we find that overall both struggle with a low recall. Empirically, we find that both BEq implementations show better recall on short statements than on long statements. Intuitively, long statements involve more logical clauses, generally making equivalence proving harder.

These results are not directly comparable to the ones presented in \citet{liu_rethinking_2024} as we do not use the same samples to evaluate. For instance, regarding their results without LLM use, which correspond to BEq$_L$, they get a recall of 67.14\% on their 200 sampled predictions, vs 30.9\% on ProofNetVerif in our case. However, given the large improvement of BEq+ over BEq$_L$ showcased in \autoref{tab:beq_performance_metrics}, it is very likely that BEq+ outperforms the original LLM-based BEq implementation. In fact, in their most compute-intensive setup, BEq achieves a recall of 72.86\% on their samples, which is only slightly better than the 67.14\% baseline. On ProofNetVerif, BEq+ outperforms BEq$_L$ with a larger relative improvement for the recall: 48.3\% vs. 30.9\%.

\subsection{\BEqP failure cases}
\label{sec:beq_failure_cases}

\subsubsection{False positives}

A typical example of two non-semantically equivalent statements that are considered equivalent by \BEqP :

\begin{lstlisting}
theorem ground_truth (a b : ℤ) :
  (ofInt a : GaussianInt) | ofInt b → a | b := sorry

theorem prediction (a b : ℤ) (ha : a | b) : a | (b : ℤ) := sorry
\end{lstlisting}

The main issue here lies from the fact that it is trivial to prove one formalization assuming the other. A generalization of this issue has been found by Simon Sorg:

\begin{lstlisting}
theorem ground_truth (n : Nat) : n + n = 2 * n := sorry

theorem prediction (p : Prop) (h : p) : p := sorry
\end{lstlisting}

A fix with minimal performance loss is available in our \href{https://github.com/augustepoiroux/LeanInteract/blob/main/examples/beq_plus.py}{implementation} and is now the default.

\subsubsection{False negatives}

False negatives with \BEqP are caused by the relatively weak power of the static proof search we use. A typical example is:

\begin{lstlisting}
theorem ground_truth : Infinite {p : Nat.Primes // p ≡ -1 [ZMOD 6]} :=

theorem prediction : Set.Infinite {p : ℕ | Nat.Prime p ∧ p % 6 = 5} :=
\end{lstlisting}

Proving the equivalence between these two statements require a consequential work as it requires to (1) prove that \lstinline{p 

\subsection{BEq+ pass@n Additional Results}
\label{sec:beq_pass_n_rlm25}

In \autoref{fig:scaling_optimal}, we conduct a study on BEq+ pass@n with 4 models: Llama3 8B, Llemma 7B, Llemma 34B, and GPT-4o on both ProofNet\# and RLM25. For ProofNet\#, we use 12-shot prompting. For RLM25, we prompt models with in-file context with proofs removed as described in \autoref{sec:real_world_results}. Llemma 34B model has not been run on the RLM25 benchmark.

\begin{figure}[t]
    \centering
    \definecolor{mygreen1}{rgb}{0.95, 0.85, 0.0}  %
\definecolor{mygreen2}{rgb}{0.7, 0.85, 0.3}  %
\definecolor{mygreen3}{rgb}{0.35, 0.55, 0.0}  %
\definecolor{mygreen4}{rgb}{0.0, 0.1, 0.0}  %

\begin{tikzpicture}
    \begin{axis}[
            name=ax1,
            width=0.28\textwidth,
            height=0.2\textheight,
            xlabel={$n$},
            ylabel={BEq+ pass@$n$ (\%)},
            xtick={1, 2, 5, 10, 20, 50},
            xticklabels={1, 2, 5, 10, 20, 50},
            x tick label style={font=\footnotesize},
            ylabel style={font=\small},
            ytick={0, 10, 20, 30},
            yticklabel style={font=\footnotesize},
            xlabel near ticks,
            ylabel near ticks,
            ymin=0, ymax=35,
            axis lines*=left,
            ymajorgrids=true,
            grid style=dashed,
            xmode=log,
            legend style={
                    at={(1.1,-0.27)},
                    anchor=north,
                    legend columns=2,
                    draw=none,
                    /tikz/every even column/.append style={column sep=0.4cm}
                },
            legend image post style={xscale=0.75}, %
        ]
        \addplot[draw=mygreen1, mark=*, mark size=1pt, thick, line width=1.2pt, mark options={fill=mygreen1, draw=mygreen1}, smooth, tension=0.3] coordinates {
                (1, 3.0) (2, 4.5) (5, 6.4) (10, 7.9) (20, 9.5) (50, 11.5)
            };

        \addplot[draw=mygreen2, mark=*, mark size=1pt, thick, line width=1.2pt, mark options={fill=mygreen2, draw=mygreen2}, smooth, tension=0.3] coordinates {
                (1, 3.5) (2, 5.1) (5, 10.5) (10, 13.8) (20, 18.7) (50, 23.0)
            };

        \addplot[draw=mygreen3, mark=*, mark size=1pt, thick, line width=1.2pt, mark options={fill=mygreen3, draw=mygreen3}, smooth, tension=0.3] coordinates {
                (1, 4.2) (2, 6.4) (5, 10.8) (10, 16.0) (20, 21.1) (50, 28.4)
            };

        \addplot[draw=mygreen4, mark=*, mark size=1pt, thick, line width=1.2pt, mark options={fill=mygreen4, draw=mygreen4}, smooth, tension=0.3] coordinates {
                (1, 13.7) (2, 18.0) (5, 21.0) (10, 23.3) (20, 26.2) (50, 30.1)
            };

        \legend{Llama3 8B, Llemma 7B, Llemma 34B, GPT-4o}
    \end{axis}

    \begin{axis}[
            at={(ax1.south east)},
            name=ax2,
            xshift=0.6cm,
            xlabel={$n$},
            width=0.28\textwidth,
            height=0.2\textheight,
            xtick={1, 2, 5, 10, 20, 50},
            xticklabels={1, 2, 5, 10, 20, 50},
            x tick label style={font=\footnotesize},
            yticklabel style={font=\footnotesize},
            ytick={10, 20, 30, 40},
            ymin=10, ymax=45,
            axis lines*=left,
            ymajorgrids=true,
            ymajorticks=true,
            grid style=dashed,
            xmode=log,
        ]
        \addplot[draw=mygreen1, mark=*, mark size=1pt, thick, line width=1.2pt, mark options={fill=mygreen1, draw=mygreen1}, smooth, tension=0.3] coordinates {
                (1, 16.22) (2, 20.85) (5, 25.06) (10, 27.83) (20, 31.30) (50, 33.84)
            };

        \addplot[draw=mygreen2, mark=*, mark size=1pt, thick, line width=1.2pt, mark options={fill=mygreen2, draw=mygreen2}, smooth, tension=0.3] coordinates {
                (1, 21.61) (2, 27.20) (5, 31.53) (10, 34.81) (20, 37.64) (50, 39.96)
            };

        \addplot[draw=mygreen3, mark=*, mark size=1pt, thick, line width=1.2pt, mark options={fill=mygreen3, draw=mygreen3}, smooth, tension=0.3] coordinates {
                (1, 25.08) (2, 29.11) (5, 34.79) (10, 37.51) (20, 41.97) (50, 44.81)
            };

        \addplot[draw=mygreen4, mark=*, mark size=1pt, thick, line width=1.2pt, mark options={fill=mygreen4, draw=mygreen4}, smooth, tension=0.3] coordinates {
                (1, 24.5) (2, 28.77) (5, 34.19) (10, 35.89) (20, 38.24) (50, 41.05)
            };

    \end{axis}

    \node[anchor=south] at (ax1.north) {\textbf{ProofNet\#}};
    \node[anchor=south] at (ax2.north) {\textbf{RLM25}};
\end{tikzpicture}
    \caption{\textbf{BEq+ pass@n scaling trends with respect to the number of samples} on ProofNet\# validation split and RLM25 using in-context learning prompting. We vary the number of candidate samples from \textit{n} = 1 to 50.}
    \label{fig:scaling_optimal}
\end{figure}

\subsection{ProofNet\#: Baseline study}
\label{sec:baseline_study}

\begin{table*}[h]
    \footnotesize
    \setlength{\tabcolsep}{3.5pt}
    \centering
    \caption{\textbf{Baseline performance on ProofNet\# using {greedy decoding}}. Except for Codex, which has been evaluated on Lean 3 in \citet{azerbayev_proofnet_2023} (indicated with an asterisk *, only the results on the test split are available), all models are evaluated on Lean 4.}
    \begin{tabular}{llcccc|cccc}
        \toprule
        \multirow{2}{*}{\textbf{Method}} & \multirow{2}{*}{\textbf{Model}} & \multicolumn{4}{c}{\textbf{Validation}} & \multicolumn{4}{c}{\textbf{Test}}                                                                                                                                                                   \\
        \cmidrule{3-10}
                                         &                                 & \textbf{Type-Check}                     & \textbf{Accuracy}$\uparrow$       & \textbf{BEq$_L$}$\uparrow$ & \textbf{BEq+}$\uparrow$ & \textbf{Type-Check} & \textbf{Accuracy}$\uparrow$ & \textbf{BEq$_L$}$\uparrow$ & \textbf{BEq+}$\uparrow$ \\
        \midrule
        12-shot                          & Codex                           & -                                       & -                                 & -                          & -                       & 23.7*               & 13.4*                       & -                          & -                       \\
        Prompt retrieval                 & Codex                           & -                                       & -                                 & -                          & -                       & 45.2*               & 16.1*                       & -                          & -                       \\
        \midrule
        MMA                              & Llama3-8B                       & 12.6                                    & 4.9                               & 2.2                        & 3.3                     & 4.3                 & -                           & 0.5                        & 0.5                     \\
        MMA                              & Llemma-7B                       & 14.2                                    & 6.0                               & 2.7                        & 4.4                     & 9.7                 & -                           & 0.0                        & 1.1                     \\
        Lean Workbook$^1$                & Llemma-7B                       & 39.5                                    & -                                 & 5.4                        & 7.0                     & 39.3                & -                           & 5.4                        & 6.5                     \\
        \midrule
        12-shot                          & Llama3-8B                       & 13.7                                    & 5.5                               & 3.3                        & 4.4                     & 13.0                & 3.3                         & 1.6                        & 3.3                     \\
        12-shot                          & Llemma-7B                       & 26.8                                    & 8.7                               & 3.3                        & 6.0                     & 29.9                & 10.9                        & 5.4                        & 6.5                     \\
        12-shot                          & Llemma-34B                      & 33.3                                    & 16.9                              & 5.5                        & 9.3                     & 29.9                & 12.5                        & 5.4                        & 7.1                     \\
        12-shot                          & GPT-4-turbo                     & 24.6                                    & 19.7                              & 8.7                        & 12.6                    & 27.7                & 22.8                        & 13.0                       & 16.8                    \\
        12-shot                          & GPT-4o                          & 33.3                                    & 26.2                              & 10.9                       & 16.4                    & 42.9                & 31.0                        & 13.6                       & 18.5                    \\
        \bottomrule
    \end{tabular}
    \label{tab:results_baseline_proofnet}
\end{table*}

We report a baseline performance in \autoref{tab:results_baseline_proofnet}. We evaluated the models described in \autoref{sec:models} using greedy decoding, coupled with either 12-shot learning or a fine-tuning on either MMA \citep{jiang_multilingual_2023} or Lean Workbook \citep{ying_lean_workbook_2024}.

\subsection{Detailed Results of Sampling-Based Methods on ProofNet\#}
\label{sec:method_detailed_results}

We present detailed results about the use of sampling-based methods on different models and autoformalization methods in \autoref{tab:results_filter_select_proofnet}.

\begin{table*}[h]
    \centering
    \setlength{\tabcolsep}{3pt}
    \small
    \caption{Evaluation results (in percentage) of sampling-based methods on ProofNet\#. For all these results, for each informal statement in the benchmark, we sampled 50 formalization attempts per model and filtered type-checking ones before applying a selection method. We observe some performance differences between the two splits which are caused by the small size of the ProofNet\# benchmark (2x 185 statements).}
    \begin{tabular}{cccccc|cccc}
        \toprule
        \multirow{2}{*}{\textbf{Model}} & \multirow{2}{*}{\makecell{\textbf{Selection}                                                                                                                                                                                                                        \\\textbf{method}}} & \multicolumn{4}{c}{\textbf{Validation}} & \multicolumn{4}{c}{\textbf{Test}} \\
        \cmidrule{3-10}
                                        &                                              & \textbf{Type-Check} & \textbf{Accuracy$\uparrow$} & \textbf{BEq$_L$$\uparrow$} & \textbf{BEq+$\uparrow$ } & \textbf{Type-Check} & \textbf{Accuracy$\uparrow$} & \textbf{BEq$_L$$\uparrow$} & \textbf{BEq+$\uparrow$} \\
        \midrule

        \multicolumn{2}{p{4cm}}{\textbf{MMA fine-tune}}                                                                                                                                                                                                                                                       \\
        \multirow{3}{*}{Llama3-8B}
                                        & Random                                       & 33.3                & \textbf{9.3}                & \textbf{3.3}               & \textbf{4.9}             & 29.0                & -                           & \textbf{2.2}               & \textbf{4.8}            \\
                                        & Majority                                     & 33.3                & 8.7                         & \textbf{3.3}               & 4.4                      & 29.0                & -                           & 1.1                        & \textbf{4.8}            \\
                                        & Self-BLEU                                    & 33.3                & 8.7                         & \textbf{3.3}               & 4.4                      & 29.0                & -                           & 1.1                        & 4.3                     \\
        \arrayrulecolor{lightgray}\hline

        \multirow{3}{*}{Llemma-7B}
                                        & Random                                       & 61.2                & 9.3                         & 4.4                        & 5.5                      & 52.2                & -                           & 3.2                        & 5.9                     \\
                                        & Majority                                     & 61.2                & 10.9                        & 3.8                        & 5.5                      & 52.2                & -                           & 3.8                        & 7.0                     \\
                                        & Self-BLEU                                    & 61.2                & \textbf{13.7}               & \textbf{4.9}               & \textbf{6.6}             & 52.2                & -                           & \textbf{5.4}               & \textbf{8.1}            \\
        \arrayrulecolor{black}\midrule

        \multicolumn{2}{p{4cm}}{\textbf{Lean Workbook fine-tune}}                                                                                                                                                                                                                                             \\
        \multirow{3}{*}{Llemma-7B}
                                        & Random                                       & 86.0                & -                           & 7.6                        & 9.2                      & 86.6                & -                           & \textbf{7.0}               & 8.6                     \\
                                        & Majority                                     & 86.0                & -                           & 9.2                        & 10.8                     & 86.6                & -                           & 5.9                        & 8.6                     \\
                                        & Self-BLEU                                    & 86.0                & -                           & \textbf{9.7}               & \textbf{12.4}            & 86.6                & -                           & 6.5                        & \textbf{10.2}           \\
        \arrayrulecolor{black}\midrule

        \multicolumn{2}{p{4cm}}{\textbf{12-shot}}                                                                                                                                                                                                                                                             \\
        \multirow{3}{*}{Llama3-8B}
                                        & Random                                       & 42.1                & 9.8                         & 4.4                        & 6.6                      & 45.7                & 13.6                        & \textbf{4.9}               & 10.3                    \\
                                        & Majority                                     & 42.1                & 12.0                        & 4.9                        & 7.1                      & 45.7                & \textbf{14.7}               & \textbf{4.9}               & \textbf{10.9}           \\
                                        & Self-BLEU                                    & 42.1                & \textbf{12.6}               & \textbf{6.0}               & \textbf{8.2}             & 45.7                & 12.0                        & \textbf{4.9}               & 9.2                     \\
        \arrayrulecolor{lightgray}\hline

        \multirow{3}{*}{Llemma-7B}
                                        & Random                                       & 84.7                & 16.9                        & 4.9                        & 7.7                      & 88.6                & 21.2                        & 9.8                        & 11.4                    \\
                                        & Majority                                     & 84.7                & 23.0                        & \textbf{8.2}               & 9.8                      & 88.6                & 23.9                        & 10.3                       & 12.0                    \\
                                        & Self-BLEU                                    & 84.7                & \textbf{23.5}               & 7.7                        & \textbf{11.5}            & 88.6                & \textbf{29.3}               & \textbf{11.4}              & \textbf{17.9}           \\
        \arrayrulecolor{lightgray}\hline

        \multirow{3}{*}{Llemma-34B}
                                        & Random                                       & 89.6                & 21.3                        & 4.9                        & 9.8                      & 84.2                & 19.6                        & 5.4                        & 11.4                    \\
                                        & Majority                                     & 89.6                & 27.3                        & \textbf{8.7}               & 12.6                     & 84.2                & 27.7                        & \textbf{10.9}              & 14.1                    \\
                                        & Self-BLEU                                    & 89.6                & \textbf{29.5}               & \textbf{8.7}               & \textbf{13.1}            & 84.2                & \textbf{28.3}               & 9.8                        & \textbf{14.7}           \\
        \arrayrulecolor{lightgray}\hline

        \multirow{3}{*}{GPT-4o}
                                        & Random                                       & 65.6                & 43.2                        & \textbf{16.4}              & \textbf{23.5}            & 70.1                & 42.9                        & 15.8                       & 22.8                    \\
                                        & Majority                                     & 65.6                & \textbf{45.4}               & 15.8                       & 21.3                     & 70.1                & 44.6                        & 15.8                       & 22.8                    \\
                                        & Self-BLEU                                    & 65.6                & 44.8                        & 15.3                       & 21.3                     & 70.1                & \textbf{45.1}               & \textbf{16.3}              & \textbf{23.4}           \\
        \arrayrulecolor{black}\bottomrule
    \end{tabular}
    \label{tab:results_filter_select_proofnet}
\end{table*}

\begin{figure}[h]
    \centering
    \definecolor{myorange1}{rgb}{1.0, 0.6, 0.2}  %
\definecolor{myorange2}{rgb}{1.0, 0.5, 0.0}  %
\definecolor{myorange3}{rgb}{0.8, 0.4, 0.0}  %
\definecolor{myorange4}{rgb}{0.6, 0.3, 0.0}  %
\definecolor{mypurple}{rgb}{0.5, 0.0, 0.5}  %
\definecolor{mybluebis}{rgb}{0.4, 0.5, 0.8}  %

\begin{tikzpicture}
    \begin{axis}[
            title={\textbf{Accuracy on ProofNet\# test split}},
            ybar=0pt,
            bar width=7pt,
            width=0.5\textwidth,
            height=0.19\textheight,
            enlarge x limits=0.2,
            ylabel={Accuracy (\%)},
            ylabel near ticks,
            symbolic x coords={Llama3 8B, Llemma 7B, Llemma 34B, GPT-4o},
            xtick=data,
            xticklabel style={font=\footnotesize, rotate=45, anchor=east},
            yticklabel style={font=\footnotesize},
            ytick={0, 10, 20, 30, 40, 50},
            legend style={
                    at={(0.5, -0.6)},
                    anchor=north,
                    legend columns=2,
                    draw=none,
                    font=\footnotesize,
                    /tikz/every even column/.append style={column sep=0.2cm}, %
                    row sep=0.05cm,
                },
            legend image post style={xscale=0.75}, %
            ymin=0, ymax=51,
            axis lines*=left,
            ymajorgrids=true,
            ymajorticks=true,
            grid style=dashed,
            area legend,
            legend cell align={left},
        ]
        \addplot[fill=mybluebis, draw opacity=0] coordinates {(Llama3 8B, 3.3) (Llemma 7B, 10.9) (Llemma 34B, 12.5) (GPT-4o, 31.0)};
        \addplot[fill=myorange1, draw opacity=0] coordinates {(Llama3 8B, 9.8) (Llemma 7B, 21.2) (Llemma 34B, 19.6) (GPT-4o, 43.0)};
        \addplot[fill=myorange3, draw opacity=0] coordinates {(Llama3 8B, 12.0) (Llemma 7B, 23.9) (Llemma 34B, 27.7) (GPT-4o, 44.6)};
        \addplot[fill=myorange4, draw opacity=0] coordinates {(Llama3 8B, 12.0) (Llemma 7B, 29.3) (Llemma 34B, 28.3) (GPT-4o, 45.1)};

        \legend{Greedy baseline, Filter + Random selection, Filter + Majority voting, Filter + Self-BLEU}
    \end{axis}
\end{tikzpicture}
    \caption{\textbf{Autoformalization accuracy on the ProofNet\# test set.} All methods use 12-shot prompting in this figure. Detailed results are reported in \autoref{tab:results_baseline_proofnet} and \autoref{tab:results_filter_select_proofnet}.}
    \label{fig:main_results_proofnet}
\end{figure}

In \autoref{fig:main_results_proofnet}, we report results on the ProofNet\# test dataset by supplementing tested models using our self-consistency method described in \autoref{sec:method}. Overall, we observe a consistent and significant improvement over the greedy baseline across all selection methods and all models evaluated. Interestingly, even using random selection over filtered generated statements is enough to outperform the greedy decoding baseline substantially. This demonstrates the practical efficacy of extensively leveraging the type-checker in the context of statement autoformalization.
We find that the overall best strategy is to use type-check filtering and Self-BLEU for the selection step. One can notice that the absolute improvement in accuracy achieved by this method ranges from +8.7\% to +18.4\% over greedy decoding, with relative improvements between 1.4x and 3x. We also report results of sampling-based methods on other classical benchmarks, PDA and MiniF2F, in \autoref{sec:minif2f_pda_results}.

\subsection{Detailed Results on RLM25}

We present detailed results about the use of sampling-based methods on different models and autoformalization methods in \autoref{tab:real_world_context_results}.

\begin{table*}[h]
    \centering
    \setlength{\tabcolsep}{2pt}
    \caption{Results on RLM25 for different methods and models. In this setup, all models are prompted with in-file context and proofs are removed. Greedy decoding is used for generation.}
    \begin{tabular}{@{}ll@{}ccc@{}}
        \toprule
        \textbf{Model} & \textbf{Method} & \textbf{Type-Check} & \textbf{BEq$_L$}  & \textbf{BEq+}     \\ \hline
        \multirow{2}{*}{Llama3 8B}
                       & No fine-tuning  & 37.19               & \underline{18.90} & \underline{20.29} \\
                       & MMA             & 41.73               & 17.55             & 19.24             \\

        \arrayrulecolor{black}\midrule
        \multirow{3}{*}{Llemma 7B}
                       & No fine-tuning  & 55.96               & \textbf{22.49}    & \underline{24.16} \\
                       & MMA             & 51.63               & 21.37             & 22.63             \\
                       & Lean Workbook   & 53.89               & 21.84             & 23.55             \\

        \arrayrulecolor{black}\midrule
        GPT-4o
                       & No fine-tuning  & 51.37               & \underline{21.18} & \textbf{24.56}    \\
        \arrayrulecolor{black}\bottomrule
    \end{tabular}
    \label{tab:real_world_context_results}
\end{table*}

\subsection{Other Benchmarks}
\label{sec:minif2f_pda_results}

While we focused our work on the ProofNet\# and RLM25 benchmarks, sampling-based methods are not specific to this benchmark, and we believe they should yield improvements out of the box on other statement autoformalization benchmarks.

In this section, we report results on two other benchmarks: Process-Driven Autoformalization (PDA) \citep{lu_process-driven_2024}, and MiniF2F \citep{zheng_minif2f_2022}.
The PDA benchmark from \citet{lu_process-driven_2024} has been designed to evaluate statement and proof autoformalization. Given the size of this benchmark, 3 test splits of 1000 theorems, and the fact that not all of these subsets contain reference formalizations, we sample 50 random problems for each test split and conduct a manual evaluation. We report our results in \autoref{tab:results_filter_select_pda} .

On all test splits, we observe that sampling-based methods largely improve over greedy decoding. We have found the real test split to be challenging as problems are sampled from Arithmo dataset \citep{jindal_2023_arithmo} without solutions, therefore requiring first solving the problem before providing an accurate formalization.

The authors of the PDA benchmark reported compilation results for statement and proof autoformalization at once. However, they do not report results for statement autoformalization alone and do not report accuracy results either. This lack of data in \citet{lu_process-driven_2024} means that we cannot compare directly to their results.

We report results on the MiniF2F benchmark in \autoref{tab:results_minif2f}. We find the Llemma 7B model fine-tuned on Lean Workbook \citep{ying_lean_workbook_2024} to perform particularly well on it. However, since the Lean Workbook dataset has been synthetically generated by a model finetuned on the MiniF2F and ProofNet benchmarks, some data leakage might have happened.

\begin{table*}[h]
    \centering
    \setlength{\tabcolsep}{3pt}
    \small
    \caption{Evaluation results (in percentage) of sampling-based methods on the \textbf{Process-Driven Autoformalization} benchmark. We evaluate on 50 samples on each of the "Basic", "Random", and "Real" splits of this benchmark. Greedy decoding is used for methods without sampling, i.e., when filtering is not mentioned. For sampling-based methods, we sample $n=50$ predictions with temperature $T=0.7$. We separate greedy decoding methods from sampling-based methods by a gray line for each model.}
    \begin{tabular}{llcc|cc|cc}
        \toprule
        \multirow{2}{*}{\textbf{Model}} & \multirow{2}{*}{\textbf{Method}} & \multicolumn{2}{c}{\textbf{Basic}} & \multicolumn{2}{c}{\textbf{Random}} & \multicolumn{2}{c}{\textbf{Real}}                                                                                   \\
        \cmidrule{3-8}
                                        &                                  & \textbf{Type-Check}                & \textbf{Accuracy$\uparrow$}         & \textbf{Type-Check}               & \textbf{Accuracy$\uparrow$} & \textbf{Type-Check} & \textbf{Accuracy$\uparrow$} \\
        \midrule
        \multirow{2}{*}{Llama3-8B}      & 12-shot                          & 18.0                               & 16.0                                & 20.0                              & 16.0                        & 20.0                & 4.0                         \\
        \arrayrulecolor{lightgray}\cline{2-8}
                                        & 12-shot + Filter + Self-BLEU     & 48.0                               & \textbf{28.0}                       & 46.0                              & \textbf{32.0}               & 72.0                & \textbf{8.0}                \\
        \arrayrulecolor{black}\hline
        \multirow{2}{*}{Llemma-7B}      & 12-shot                          & 20.0                               & 14.0                                & 30.0                              & 26.0                        & 62.0                & 0.0                         \\
        \arrayrulecolor{lightgray}\cline{2-8}
                                        & 12-shot + Filter + Self-BLEU     & 76.0                               & \textbf{58.0}                       & 72.0                              & \textbf{54.0}               & 100.0               & \textbf{8.0}                \\
        \arrayrulecolor{black}\hline
        \multirow{2}{*}{GPT-4o}         & 12-shot                          & 30.0                               & 28.0                                & 42.0                              & 38.0                        & 4.0                 & 0.0                         \\
        \arrayrulecolor{lightgray}\cline{2-8}
                                        & 12-shot + Filter + Self-BLEU     & 64.0                               & \textbf{54.0}                       & 66.0                              & \textbf{56.0}               & 26.0                & \textbf{12.0}               \\
        \arrayrulecolor{black}\bottomrule
    \end{tabular}
    \label{tab:results_filter_select_pda}
\end{table*}

\begin{table*}[h]
    \footnotesize
    \setlength{\tabcolsep}{3pt}
    \centering
    \caption{Performance of different methods on \textbf{MiniF2F}. Greedy decoding is used for methods without sampling, i.e., when filtering is not mentioned. For sampling-based methods, we sample $n=50$ predictions with temperature $T=0.7$. We separate greedy decoding methods from sampling-based methods by a gray line for each model. $^1$Lean Workbook dataset has been curated with a model trained on the MiniF2F benchmark, thus data leakage concerns apply.}
    \begin{tabular}{llccc}
        \toprule
        \textbf{Model} & \textbf{Method}                        & \textbf{Type-Check} & \textbf{BEq$_L$}$\uparrow$ & \textbf{BEq+}$\uparrow$ \\
        \midrule

        \multirow{3}{*}{Llama3 8B}
                       & 12-shot                                & 45.9                & 6.8                        & 14.6                    \\ %
                       & MMA                                    & 36.7                & 3.1                        & 8.8                     \\
        \arrayrulecolor{lightgray}\cline{2-5}
                       & 12-shot + Filter + Self-BLEU           & 93.0                & 10.3                       & 21.3                    \\
        \arrayrulecolor{black}\midrule

        \multirow{5}{*}{Llemma 7B}
                       & 12-shot                                & 67.0                & 7.6                        & 16.0                    \\
                       & MMA                                    & 32.2                & 2.1                        & 5.3                     \\
                       & Lean Workbook$^1$                      & 89.6                & 14.3                       & 28.5                    \\ %
        \arrayrulecolor{lightgray}\cline{2-5}
                       & 12-shot + Filter + Self-BLEU           & 99.8                & 10.3                       & 21.3                    \\ %
                       & Lean Workbook + Filter + Self-BLEU$^1$ & 99.0                & 14.1                       & 28.1                    \\ %
        \arrayrulecolor{black}\midrule

        GPT-4o         & 12-shot                                & 24.4                & 8.0                        & 13.5                    \\
        \arrayrulecolor{black}\bottomrule
    \end{tabular}
    \vspace{0.5em}
    \label{tab:results_minif2f}
\end{table*}

\subsection{Low-correction Effort Formalizations}
\label{sec:accuracy_low_corr_effort}

\begin{figure}[h]
    \centering
    \definecolor{myorange1}{rgb}{1.0, 0.6, 0.2}  %
\definecolor{myorange2}{rgb}{1.0, 0.5, 0.0}  %
\definecolor{myorange3}{rgb}{0.8, 0.4, 0.0}  %
\definecolor{myorange4}{rgb}{0.6, 0.3, 0.0}  %
\definecolor{mypurple}{rgb}{0.5, 0.0, 0.5}  %
\definecolor{mybluebis}{rgb}{0.4, 0.5, 0.8}  %

\begin{tikzpicture}
    
\begin{axis}[
        title={\textbf{Accuracy - Low correction effort}},
        ybar=0pt,
        bar width=7pt,
        xshift=0.2cm,
        width=0.5\textwidth,
        height=0.19\textheight,
        enlarge x limits=0.2,
        ylabel={Accuracy (\%)},
        symbolic x coords={Llama3 8B, Llemma 7B, Llemma 34B, GPT-4o},
        xtick=data,
        xticklabel style={font=\footnotesize, rotate=45, anchor=east},
        ylabel style={font=\small},
        yticklabel style={font=\small},
        ytick={0, 10, 20, 30, 40, 50, 60},
        yticklabels={0,, 20,, 40,, 60},
        legend style={
            at={(0.45, -0.6)},
            anchor=north,
            legend columns=2,
            draw=none,
            font=\footnotesize,
            /tikz/every even column/.append style={column sep=0.2cm}, %
            row sep=0.05cm,
        },
        ymin=0, ymax=62,
        axis lines*=left,
        ymajorgrids=true,
        ymajorticks=true,
        grid style=dashed,
        area legend,
        legend cell align={left},
    ]
        \addplot[fill=mybluebis, draw opacity=0] coordinates {(Llama3 8B, 9.1) (Llemma 7B, 16.7) (Llemma 34B, 19.9) (GPT-4o, 40.3)};
        \addplot[fill=myorange1, draw opacity=0] coordinates {(Llama3 8B, 23.7) (Llemma 7B, 37.1) (Llemma 34B, 35.5) (GPT-4o, 60.7)};
        \addplot[fill=myorange3, draw opacity=0] coordinates {(Llama3 8B, 25.8) (Llemma 7B, 40.9) (Llemma 34B, 40.3) (GPT-4o, 60.2)};
        \addplot[fill=myorange4, draw opacity=0] coordinates {(Llama3 8B, 25.3) (Llemma 7B, 48.9) (Llemma 34B, 51.1) (GPT-4o, 61.3)};
        
        \legend{Greedy baseline, Filter + Random, Filter + Majority voting, Filter + Self-BLEU}
\end{axis}
\end{tikzpicture}
    \caption{\textbf{Proportion of formalized statements evaluated as correct or as fixable with a low amount of effort (i.e., \textit{close-to-correct}) on the ProofNet\# test set.} All models are prompted with 12-shot examples. Detailed results are reported in \autoref{tab:results_fixable}.}
    \label{fig:fixable_results_proofnet}
\end{figure}

One goal of autoformalization is the development of AI-assisted tools for formalization. In this setting, producing close-to-correct formal statements can already help users by providing hints and potential directions. Using the same setup as in the previous section, we report our results on the ProofNet\# test split in \autoref{fig:main_results_proofnet}. Note that in this setup, contrary to \citet{jiang_multilingual_2023}, we are considering only well-typed statements.

We find that, by using sampling-based methods, open-source models Llemma-7B and Llemma-34B can autoformalize $\sim 50\%$ of the mathematical statements from the ProofNet\# test benchmark in a \textit{close-to-correct} way\footnote{We define \textit{close-to-correct} formalizations as those with one slightly diverging hypothesis or conclusion, typically fixable in a matter of seconds.}. This makes these open models good fit for local autoformalization assistant, especially Llemma-7B for its relatively small size.

In this section, we present several examples of autoformalizations on ProofNet\# validation split that are evaluated as incorrect yet fixable with low effort. Evaluation results on ProofNet\# test split are presented in \autoref{tab:results_fixable}.

\begin{table}[h]
    \centering
    \caption{Models performance (in percentage) on ProofNet\# test split when accounting for formalizations that can be \textbf{corrected with a low amount of efforts}.}
    \small
    \begin{tabular}{p{2.0cm}p{3cm}c}
        \toprule
        Model                       & Method             & Accuracy$\uparrow$ \\
        \midrule
        \multirow{4}{*}{Llama3-8B}  & Greedy             & 9.1                \\
                                    & Filter + Random    & 23.7               \\
                                    & Filter + Majority  & \textbf{25.8}      \\
                                    & Filter + Self-BLEU & 25.3               \\
        \arrayrulecolor{lightgray}\hline
        \multirow{4}{*}{Llemma-7B}  & Greedy             & 16.7               \\
                                    & Filter + Random    & 37.1               \\
                                    & Filter + Majority  & 40.9               \\
                                    & Filter + Self-BLEU & \textbf{48.9}      \\
        \arrayrulecolor{lightgray}\hline
        \multirow{4}{*}{Llemma-34B} & Greedy             & 19.9               \\
                                    & Filter + Random    & 35.5               \\
                                    & Filter + Majority  & 40.3               \\
                                    & Filter + Self-BLEU & \textbf{51.1}      \\
        \arrayrulecolor{lightgray}\hline
        \multirow{4}{*}{GPT-4o}     & Greedy             & 40.3               \\
                                    & Filter + Random    & 60.7               \\
                                    & Filter + Majority  & 60.2               \\
                                    & Filter + Self-BLEU & \textbf{61.3}      \\
        \arrayrulecolor{black}\bottomrule
    \end{tabular}
    \label{tab:results_fixable}
\end{table}

\begin{tcolorbox}[title=Low-correction effort examples]
    \subsubsection*{Rudin exercise\_1\_1a}

    \textbf{Informal statement:} If $r$ is rational $(r \neq 0)$ and $x$ is irrational, prove that $r+x$ is irrational.

    \textbf{Generated formalization:}
    \begin{lstlisting}
theorem exercise_1_1a (hr : r ≠ 0) (hi : Irrational x) : Irrational (r + x) :=
\end{lstlisting}

    \textbf{Issue:} $r$ has not been declared as a rational number.

    \noindent\textcolor{lightgray}{\rule{\linewidth}{0.1mm}}

    \subsubsection*{Herstein exercise\_4\_3\_25}

    \textbf{Informal statement:} Let $R$ be the ring of $2 \times 2$ matrices over the real numbers; suppose that $I$ is an ideal of $R$. Show that $I = (0)$ or $I=R$.

    \textbf{Generated formalization:}
    \begin{lstlisting}
theorem exercise_4_3_25 {R : Type*} [CommRing R] (I : Ideal (Matrix (Fin 2) (Fin 2) ℝ)) : I = ⊥ ∨ I = ⊤ :=
\end{lstlisting}

    \textbf{Issue:} Superfluous declaration of \lstinline{CommRing R}.

\end{tcolorbox}

\subsection{Ablation Study: Detailed Results}

We present detailed results of our ablation study on the type-check filtering step in \autoref{tab:results_ablation_study}.

\begin{table*}[h]
    \centering
    \caption{Models performance (in percentage) on ProofNet\# validation split removing different aspects of sampling-based methods. We also report Greedy baseline and the Filter + Self-BLEU results as reference.}
    \begin{tabular}{p{2.5cm}p{3.5cm}ccccc}
        \toprule
        Model                       & Method                & Type-Check & Accuracy$\uparrow$ & BEq$_L$$\uparrow$ & BEq+$\uparrow$ \\
        \midrule
        \multirow{6}{*}{Llama3-8B}  & Greedy                & 13.7       & 5.5                & 3.3               & 4.4            \\
                                    & No filter + Random    & 9.8        & 4.4                & 2.7               & 2.7            \\
                                    & No filter + Majority  & 13.1       & 5.5                & 3.3               & 3.8            \\
                                    & No filter + Self-BLEU & 14.2       & 6.0                & 2.7               & 3.8            \\
                                    & Filter + Random       & 42.1       & 9.8                & 4.4               & 6.6            \\
                                    & Filter + Self-BLEU    & 42.1       & \textbf{12.6}      & \textbf{6.0}      & \textbf{8.2}   \\
        \arrayrulecolor{lightgray}\hline
        \multirow{6}{*}{Llemma-7B}  & Greedy                & 26.8       & 8.7                & 3.3               & 6.0            \\
                                    & No filter + Random    & 25.7       & 6.0                & 3.3               & 3.8            \\
                                    & No filter + Majority  & 25.1       & 10.9               & 4.4               & 4.9            \\
                                    & No filter + Self-BLEU & 32.2       & 14.2               & 6.6               & 9.3            \\
                                    & Filter + Random       & 84.7       & 16.9               & 4.9               & 7.7            \\
                                    & Filter + Self-BLEU    & 84.7       & \textbf{23.5}      & \textbf{7.7}      & \textbf{11.5}  \\
        \arrayrulecolor{lightgray}\hline
        \multirow{6}{*}{Llemma-34B} & Greedy                & 33.3       & 16.9               & 5.5               & 9.3            \\
                                    & No filter + Random    & 24.6       & 9.3                & 2.7               & 4.4            \\
                                    & No filter + Majority  & 24.6       & 10.9               & 4.9               & 7.1            \\
                                    & No filter + Self-BLEU & 32.8       & 14.2               & 3.8               & 6.0            \\
                                    & Filter + Random       & 89.6       & 21.3               & 4.9               & 9.8            \\
                                    & Filter + Self-BLEU    & 89.6       & \textbf{29.5}      & \textbf{8.7}      & \textbf{13.1}  \\
        \arrayrulecolor{lightgray}\hline
        \multirow{6}{*}{GPT-4o}     & Greedy                & 33.3       & 26.2               & 10.9              & 16.4           \\
                                    & No filter + Random    & 33.3       & 25.7               & 9.8               & 14.8           \\
                                    & No filter + Majority  & 35.5       & 29.5               & 12.0              & 16.4           \\
                                    & No filter + Self-BLEU & 36.1       & 28.4               & 12.0              & 16.9           \\
                                    & Filter + Random       & 65.6       & 43.2               & \textbf{16.4}     & \textbf{23.5}  \\
                                    & Filter + Self-BLEU    & 65.6       & \textbf{44.8}      & 15.3              & 21.3           \\
        \arrayrulecolor{black}\bottomrule
    \end{tabular}
    \label{tab:results_ablation_study}
\end{table*}

\subsection{Sampling Scaling}

Detailed results about our sampling study can be found in \autoref{tab:results_sampling_scaling}.

\begin{table*}[h]
    \small
    \setlength{\tabcolsep}{3pt}
    \centering
    \caption{Evaluation results (in percentage) of sampling-based methods on ProofNet\# validation split for different numbers of formalizations sampled during the sampling phase (represented by the number $n$ in this table). We used a 12-shot prompt with the filter+Self-BLEU variant and a temperature of $0.7$.}
    \hspace*{-0.5cm}
    \begin{tabular}{ccccc|cccc|cccc|cccc}
        \multirow{2}{*}{\textbf{Model}} & \multicolumn{4}{c}{\textbf{Type-Check}} & \multicolumn{4}{c}{\textbf{Accuracy$\uparrow$}} & \multicolumn{4}{c}{\textbf{BEq$_L$$\uparrow$}} & \multicolumn{4}{c}{\textbf{BEq+$\uparrow$}}                                                                                    \\
        \cmidrule{2-17}
                                        & n=1                                     & n=5                                             & n=20                                           & n=50                                        & n=1  & n=5  & n=20 & n=50 & n=1 & n=5  & n=20 & n=50 & n=1  & n=5  & n=20 & n=50 \\
        \midrule
        Llama3-8B                       & 9.8                                     & 22.4                                            & 32.8                                           & 42.1                                        & 4.4  & 7.1  & 10.4 & 12.6 & 2.7 & 3.3  & 4.4  & 6.0  & 2.7  & 4.9  & 6.0  & 8.2  \\
        Llemma-7B                       & 25.7                                    & 50.3                                            & 71.0                                           & 84.7                                        & 6.0  & 10.9 & 19.7 & 23.5 & 3.3 & 3.3  & 7.1  & 7.7  & 3.8  & 6.0  & 10.4 & 11.5 \\
        Llemma-34B                      & 24.6                                    & 55.2                                            & 78.7                                           & 89.6                                        & 9.3  & 14.8 & 25.7 & 29.5 & 2.7 & 5.5  & 8.2  & 8.7  & 4.4  & 8.7  & 12.0 & 13.1 \\
        GPT-4o                          & 33.3                                    & 47.0                                            & 55.7                                           & 65.6                                        & 25.7 & 34.4 & 38.8 & 44.8 & 9.8 & 14.2 & 13.7 & 15.3 & 14.8 & 20.2 & 19.7 & 21.3 \\
        \arrayrulecolor{black}%
    \end{tabular}
    \label{tab:results_sampling_scaling}
\end{table*}

\clearpage

\subsection{12-shot Prompt}
\label{sec:few_shot_prompt}

\textbf{Note:} We translated the 12-shot prompt from ProofNet\# to Lean 4, with as minimal changes as possible, for the accuracy comparison with previous results to be as fair as possible. In particular, we did not remove/change the statements leaked from the benchmark and did not correct potential formalization mistakes in this prompt to make our results comparable with the results in \citet{azerbayev_proofnet_2023}.

\begin{tcolorbox}[title=12-shot examples]
    Natural language version:

    Let $P$ be a $p$-subgroup of $G$. Then $P$ is contained in a Sylow $p$-subgroup of $G$.

    Translate the natural language version to a Lean 4 version:
    \begin{lstlisting}
theorem exists_le_sylow [Group G] {P : Subgroup G} (hP : IsPGroup p P) : ∃ Q : Sylow p G, P ≤ Q :=
\end{lstlisting}

    \noindent\textcolor{lightgray}{\rule{\linewidth}{0.1mm}}

    Natural language version:

    Let $E$ and $F$ be complex normed spaces and let $f:E\to F$. If $f$ is differentiable and bounded, then $f$ is constant
    Translate the natural language version to a Lean 4 version:

    \begin{lstlisting}
theorem exists_eq_const_of_bounded {E : Type u} [NormedAddCommGroup E] [NormedSpace ℂ E] {F : Type v} [NormedAddCommGroup F] [NormedSpace ℂ F] {f : E → F} (hf : Differentiable ℂ f)(hb : IsBounded (range f)) : ∃ c, f = const E c :=
\end{lstlisting}

    \noindent\textcolor{lightgray}{\rule{\linewidth}{0.1mm}}

    Natural language version:

    Let $X$ be a topological space; let $A$ be a subset of $X$. Suppose that for each $x\in A$ there is an open set $U$ containing $x$ such that $U\subset A$. Then $A$ is open in $X$.

    Translate the natural language version to a Lean 4 version:
    \begin{lstlisting}
theorem subset_of_open_subset_is_open (X : Type*) [TopologicalSpace X]
(A : Set X) (hA : ∀ x ∈ A, ∃ U : Set X, IsOpen U ∧ x ∈ U ∧ U ⊆ A):
  IsOpen A :=
\end{lstlisting}

    \noindent\textcolor{lightgray}{\rule{\linewidth}{0.1mm}}

    Natural language version:

    Two multiplicative functions $f, g:\mathbb{N}\to R$ are equal if and only if $f(p^i)=f(g^i)$ for all primes $p$.

    Translate the natural language version to a Lean 4 version:
    \begin{lstlisting}
theorem eq_iff_eq_on_prime_powers [CommMonoidWithZero R] (f : ArithmeticFunction R)
(hf : f.IsMultiplicative) (g : ArithmeticFunction R) (hg : g.IsMultiplicative) :
f = g ↔ ∀ p i : ℕ, Nat.Prime p → f (p ^ i) = g (p ^ i) :=
\end{lstlisting}

    \noindent\textcolor{lightgray}{\rule{\linewidth}{0.1mm}}

    Natural language version:

    If $z_1, \dots, z_n$ are complex, then $|z_1 + z_2 + \dots + z_n|\leq |z_1| + |z_2| + \dots + |z_n|$.

    Translate the natural language version to a Lean 4 version:
    \begin{lstlisting}
theorem abs_sum_leq_sum_abs (n : ℕ) (f : ℕ → ℂ) :
abs (∑ i in Finset.range n, f i) ≤ ∑ i in Finset.range n, abs (f i) :=
\end{lstlisting}

    \noindent\textcolor{lightgray}{\rule{\linewidth}{0.1mm}}

    Natural language version:

    If x and y are in $\mathbb{R}^n$, then $|x+y|^2 + |x-y|^2 = 2|x|^2 + 2|y|^2$.

    Translate the natural language version to a Lean 4 version:
    \begin{lstlisting}
theorem sum_add_square_sub_square_eq_sum_square (n : ℕ) (x y : EuclideanSpace ℝ (Fin n)) :
‖x + y‖^2 + ‖x - y‖^2 = 2*‖x‖^2 + 2*‖y‖^2 :=
\end{lstlisting}

    \noindent\textcolor{lightgray}{\rule{\linewidth}{0.1mm}}

    Natural language version:

    If $x$ is an element of infinite order in $G$, prove that the elements $x^n$, $n \in \mathbb{Z}$ are all distinct.

    Translate the natural language version to a Lean 4 version:
    \begin{lstlisting}
theorem distinct_powers_of_infinite_order_element (G : Type*) [Group G] (x : G)
  (hx_inf : ∀ n : ℕ, x ^ n ≠ 1) :
  ∀ m n : ℤ, m ≠ n → x ^ m ≠ x ^ n :=
\end{lstlisting}

    \noindent\textcolor{lightgray}{\rule{\linewidth}{0.1mm}}

    Natural language version:

    A set of vectors $\{v_i\}_{i\in I}$ orthogonal with respect to some bilinear form $B: V\times V\to K$ is linearly independent if for all $i \in I, B(v_i, v_i)\neq 0$.

    Translate the natural language version to a Lean 4 version:
    \begin{lstlisting}
theorem linear_independent_of_is_Ortho {V K : Type*} [Field K]
[AddCommGroup V] [Module K V] {n : Type*} {B : BilinForm K V}
{v : n → V} (hv₁ : B.iIsOrtho v)
(hv₂ : ∀ (i : n), ¬B.IsOrtho (v i) (v i)) :
LinearIndependent K v :=
\end{lstlisting}

    \noindent\textcolor{lightgray}{\rule{\linewidth}{0.1mm}}

    Natural language version:

    Suppose that $V$ is an $n$-dimensional vector space. Then for some set of vectors $\{v_i\}_{i=1}^k$, if $k>n$ then there exist scalars $f_1, \dots, f_k$ such that $\sum_{i=1}^k f_kv_k = 0$.

    Translate the natural language version to a Lean 4 version:
    \begin{lstlisting}
theorem exists_nontrivial_relation_sum_zero_of_
  dim_succ_lt_card {K V : Type*}
[DivisionRing K] [AddCommGroup V] [Module K V] [FiniteDimensional K V]
{t : Finset V} (h : FiniteDimensional.finrank K V + 1 < t.card) :
∃ (f : V → K), t.sum (λ (e : V) => f e • e) = 0 ∧ t.sum (λ (e : V) => f e) = 0
∧ ∃ (x : V) (H : x ∈ t), f x ≠ 0 :=
\end{lstlisting}

    \noindent\textcolor{lightgray}{\rule{\linewidth}{0.1mm}}

    Natural language version:

    A group is commutative if the quotient by the center is cyclic.

    Translate the natural language version to a Lean 4 version:
    \begin{lstlisting}
theorem comm_group_of_cycle_center_quotient {G H : Type*} [Group G] [Group H]
[IsCyclic H] (f : G →* H) (hf : f.ker ≤ (center G : Subgroup G)):
CommGroup G :=
\end{lstlisting}

    \noindent\textcolor{lightgray}{\rule{\linewidth}{0.1mm}}

    Natural language version:

    If $H$ is a $p$-subgroup of $G$, then the index of $H$ inside its normalizer is congruent modulo $p$ to the index of $H$.

    Translate the natural language version to a Lean 4 version:
    \begin{lstlisting}
theorem card_quotient_normalizer_modEq_
  card_quotient {G : Type*} [Group G] [Fintype G] {p : ℕ} {n : ℕ} [hp : Fact p.Prime]
{H : Subgroup G} (hH : Fintype.card H = p ^ n) :
Fintype.card (normalizer H / Subgroup.comap ((normalizer H).subtype : normalizer H →* G) H) ≡
  Fintype.card (G / H) [MOD p] :=
\end{lstlisting}

    \noindent\textcolor{lightgray}{\rule{\linewidth}{0.1mm}}

    Natural language version:

    Suppose $X, Y, Z$ are metric spaces, and $Y$ is compact. Let $f$ map $X$ into $Y$, let $g$ be a continuous one-to-one mapping of $Y$ into $Z$, and put $h(x)=g(f(x))$ for $x \in X$. Prove that $f$ is uniformly continuous if $h$ is uniformly continuous.

    Translate the natural language version to a Lean 4 version:
    \begin{lstlisting}
theorem uniform_continuous_of_continuous_
   injective_uniform_continuous_comp
{X Y Z : Type*} [MetricSpace X] [MetricSpace Y] [MetricSpace Z]
(hY : CompactSpace Y) (f : X → Y) (g : Y → Z) (hgc : Continuous g)
(hgi : Function.Injective g)
(h : UniformContinuous (g ∘ f)) : UniformContinuous f :=
\end{lstlisting}
\end{tcolorbox}

\subsection{Data Contamination}
\label{sec:appendix_contamination}

Data contamination is a serious issue in today's LLM benchmarks. In fact, large language models are trained on large-scale training data, thus, despite the filtering efforts, data leakage might happen. For the new dataset RLM25 we introduce, as stated in \autoref{sec:rlm25_description}, all projects selected for evaluation were made available after the knowledge cutoff dates of the evaluated models. In particular, Llemma 7B, performing almost on par with GPT-4o on this benchmark, is open-weight and has been released in August 2023, thus before the first commit of any of these projects.

ProofNet 3 was released in February 2023, and an unofficial port to Lean 4 has been publicly available since March 2024. Since the cutoff training dates for all models used in these experiments are before March 2024, Lean 4 data contamination due to training is not possible. However, it remains theoretically possible that some models were trained on the Lean 3 version and weakly generalized to Lean 4.
Such data leakage for the Llemma models family \citep{azerbayev_llemma_2023} seems unlikely as the authors claim they have specifically excluded ProofNet from their training data.

For our data contamination study, we use an unofficial Lean 4 port \citep{vishwakarma_rahul3613proofnet-lean4_2024} of the ProofNet benchmark made by an independent research team. This port shows minimal differences from the original Lean 3 ProofNet benchmark, preserving the order of hypotheses and terms. Upon analyzing the raw predictions of all models, we did not find any exact matches with the Lean 4 ground truths. This is primarily because the theorems in the benchmark follow an \lstinline{exercise_number} naming scheme, which the models do not produce. Consequently, we employed fuzzy matching for our data contamination checks. This involved normalizing whitespaces and removing comments and theorem names. We found a maximum of 2.2\% matches (4 statements out of 185/186) for each model independently on the validation split, including the 2 statements leaked by the prompt. Given that the space of correct formal statements is heavily constrained, this hit rate is quite reasonable.
Below, we provide a list of all unique hits found across all models and experiments.
Most of these hits are very short and almost unavoidable. Considering these results, it seems unlikely that significant data leakage occurred during the training of these models.

Nonetheless, during our data contamination study, we found that 4 examples from the 12-shot prompt in \citet{azerbayev_proofnet_2023}, which we intended to compare to, were also present in the benchmark (2 in the validation set and 2 in the test set). Fortunately, this affects the results only negligibly (at most $\sim 1.1\%$). We report all our results with these statements removed.

\begin{tcolorbox}[title=List of all the hits found (using fuzzy matching) across all our experiments on the ProofNet validation split]
    \textbf{Munkres|exercise\_29\_1:}
    \begin{lstlisting}
theorem exercise_29_1 : ¬ LocallyCompactSpace ℚ :=
\end{lstlisting}

    \noindent\textcolor{lightgray}{\rule{\linewidth}{0.1mm}}

    \textbf{Dummit-Foote|exercise\_1\_1\_22a:}
    \begin{lstlisting}
theorem exercise_1_1_22a {G : Type*} [Group G] (x g : G) :
  orderOf x = orderOf (g⁻¹ * x * g) :=
\end{lstlisting}

    \noindent\textcolor{lightgray}{\rule{\linewidth}{0.1mm}}

    \textbf{Herstein|exercise\_2\_1\_27:}
    \begin{lstlisting}
theorem exercise_2_1_27 {G : Type*} [Group G]
  [Fintype G] : ∃ (m : ℕ), ∀ (a : G), a ^ m = 1 :=
\end{lstlisting}

    \noindent\textcolor{lightgray}{\rule{\linewidth}{0.1mm}}

    \textbf{Munkres|exercise\_17\_4:}
    \begin{lstlisting}
theorem exercise_17_4 {X : Type*} [TopologicalSpace X]
  (U A : Set X) (hU : IsOpen U) (hA : IsClosed A) :
  IsOpen (U \ A) ∧ IsClosed (A \ U) :=
\end{lstlisting}

    \noindent\textcolor{lightgray}{\rule{\linewidth}{0.1mm}}

    \textbf{Herstein|exercise\_5\_5\_2:}
    \begin{lstlisting}
theorem exercise_5_5_2 : Irreducible (X^3 - 3*X - 1 : Polynomial ℚ) :=
\end{lstlisting}

    \noindent\textcolor{lightgray}{\rule{\linewidth}{0.1mm}}

    \textbf{Munkres|exercise\_32\_3:}
    \begin{lstlisting}
theorem exercise_32_3 {X : Type*} [TopologicalSpace X]
  (hX : LocallyCompactSpace X) (hX' : T2Space X) :
  RegularSpace X :=
\end{lstlisting}

    \noindent\textcolor{lightgray}{\rule{\linewidth}{0.1mm}}

    \textbf{Herstein|exercise\_4\_3\_25:}
    \begin{lstlisting}
theorem exercise_4_3_25 (I : Ideal (Matrix (Fin 2) (Fin 2) ℝ)) :
  I = ⊥ ∨ I = ⊤ :=
\end{lstlisting}
\end{tcolorbox}

\begin{tcolorbox}[title=Statements from the validation split leaked by the ProofNet prompt]
    \textbf{Munkres|exercise\_13\_1}
    \begin{lstlisting}
theorem subset_of_open_subset_is_open (X : Type*) [TopologicalSpace X]
  (A : Set X) (hA : ∀ x ∈ A, ∃ U : Set X, IsOpen U ∧ x ∈ U ∧ U ⊆ A):
    IsOpen A :=
\end{lstlisting}

    \noindent\textcolor{lightgray}{\rule{\linewidth}{0.1mm}}

    \textbf{Dummit-Foot|exercise\_1\_1\_34}
    \begin{lstlisting}
theorem distinct_powers_of_infinite_order_element (G : Type*) [Group G] (x : G)
  (hx_inf : ∀ n : ℕ, x ^ n ≠ 1) :
  ∀ m n : ℤ, m ≠ n → x ^ m ≠ x ^ n :=
\end{lstlisting}
\end{tcolorbox}

\onecolumn
\subsection{All results on ProofNet\#}
\label{sec:all_results_proofnet}

We report in the table below metric results for all autoformalization methods and models on which we conducted manual evaluation. Such manual evaluation has been conducted on both ProofNet\# validation and test splits.

\scriptsize
\begin{longtable}{@{}lllrrrrr@{}}
    \toprule
    Model                   & Strategy                           & Split & Type Check & Accuracy & \BEqL & \BEqP & Hyp. Rej. \\
    \midrule
    ensemble-12shot-top50   & top50 type-checked majority voting & test  & 96.2       & 44.6     & 17.4  & 23.4  & 3.8       \\
    ensemble-12shot-top50   & top50 type-checked random          & test  & 96.2       & 33.2     & 9.8   & 14.7  & 2.2       \\
    ensemble-12shot-top50   & top50 type-checked self consistent & test  & 96.2       & 48.4     & 17.9  & 26.6  & 3.3       \\
    gpt-4-turbo-2024-04-09  & greedy type-checked                & test  & 27.7       & 22.8     & 13.0  & 16.8  & 1.1       \\
    gpt-4-turbo-2024-04-09  & greedy type-checked                & valid & 24.6       & 19.7     & 8.7   & 12.6  & 0.0       \\
    gpt-4o-2024-05-13       & greedy type-checked                & test  & 42.9       & 31.0     & 13.6  & 18.5  & 1.6       \\
    gpt-4o-2024-05-13       & greedy type-checked                & valid & 33.3       & 26.2     & 10.9  & 16.4  & 0.0       \\
    gpt-4o-2024-05-13-top50 & top50 type-checked majority voting & test  & 70.1       & 44.6     & 15.8  & 22.8  & 2.2       \\
    gpt-4o-2024-05-13-top50 & top50 type-checked random          & test  & 70.1       & 42.9     & 15.8  & 22.8  & 1.1       \\
    gpt-4o-2024-05-13-top50 & top50 type-checked self consistent & test  & 70.1       & 45.1     & 16.3  & 23.4  & 2.2       \\
    gpt-4o-2024-05-13-top50 & top20 type-checked self consistent & valid & 55.7       & 38.8     & 13.7  & 19.7  & 0.5       \\
    gpt-4o-2024-05-13-top50 & top50 majority voting type-checked & valid & 35.5       & 29.5     & 12.0  & 16.4  & 0.0       \\
    gpt-4o-2024-05-13-top50 & top50 random type-checked          & valid & 33.3       & 25.7     & 9.8   & 14.8  & 0.0       \\
    gpt-4o-2024-05-13-top50 & top50 self consistent type-checked & valid & 36.1       & 28.4     & 12.0  & 16.9  & 0.0       \\
    gpt-4o-2024-05-13-top50 & top50 type-checked majority voting & valid & 65.6       & 45.4     & 15.8  & 21.3  & 0.0       \\
    gpt-4o-2024-05-13-top50 & top50 type-checked random          & valid & 65.6       & 43.2     & 16.4  & 23.5  & 0.0       \\
    gpt-4o-2024-05-13-top50 & top50 type-checked self consistent & valid & 65.6       & 44.8     & 15.3  & 21.3  & 0.0       \\
    gpt-4o-2024-05-13-top50 & top5 type-checked self consistent  & valid & 47.0       & 34.4     & 14.2  & 20.2  & 0.0       \\
    llama3-8b               & greedy type-checked                & test  & 13.0       & 3.3      & 1.6   & 3.3   & 1.1       \\
    llama3-8b               & greedy type-checked                & valid & 13.7       & 5.5      & 3.3   & 4.4   & 0.5       \\
    llama3-8b-mma           & valid                              & valid & 12.6       & 4.9      & 2.2   & 3.3   & 2.2       \\
    llama3-8b-mma-top50     & top50 type-checked most frequent   & valid & 33.3       & 8.7      & 3.3   & 4.4   & 2.2       \\
    llama3-8b-mma-top50     & top50 type-checked random          & valid & 33.3       & 9.3      & 3.3   & 4.9   & 2.7       \\
    llama3-8b-mma-top50     & top50 type-checked self consistent & valid & 33.3       & 8.7      & 3.3   & 4.4   & 2.7       \\
    llama3-8b-top50         & top50 type-checked majority voting & test  & 45.7       & 14.7     & 4.9   & 10.9  & 2.2       \\
    llama3-8b-top50         & top50 type-checked random          & test  & 45.7       & 13.6     & 4.9   & 10.3  & 1.1       \\
    llama3-8b-top50         & top50 type-checked self consistent & test  & 45.7       & 12.0     & 4.9   & 9.2   & 1.6       \\
    llama3-8b-top50         & top20 type-checked self consistent & valid & 32.8       & 10.4     & 4.4   & 6.0   & 0.5       \\
    llama3-8b-top50         & top50 majority voting type-checked & valid & 13.1       & 5.5      & 3.3   & 3.8   & 0.5       \\
    llama3-8b-top50         & top50 random type-checked          & valid & 9.8        & 4.4      & 2.7   & 2.7   & 0.5       \\
    llama3-8b-top50         & top50 self consistent type-checked & valid & 14.2       & 6.0      & 2.7   & 3.8   & 0.5       \\
    llama3-8b-top50         & top50 type-checked most frequent   & valid & 42.1       & 12.0     & 4.9   & 7.1   & 1.1       \\
    llama3-8b-top50         & top50 type-checked random          & valid & 42.1       & 9.8      & 4.4   & 6.6   & 1.1       \\
    llama3-8b-top50         & top50 type-checked self consistent & valid & 42.1       & 12.6     & 6.0   & 8.2   & 1.1       \\
    llama3-8b-top50         & top5 type-checked self consistent  & valid & 22.4       & 7.1      & 3.3   & 4.9   & 1.1       \\
    llemma-34b              & greedy type-checked                & test  & 29.9       & 12.5     & 5.4   & 7.1   & 1.1       \\
    llemma-34b              & greedy type-checked                & valid & 33.3       & 16.9     & 5.5   & 9.3   & 0.0       \\
    llemma-34b-top50        & top50 type-checked majority voting & test  & 84.2       & 27.7     & 10.9  & 14.1  & 3.3       \\
    llemma-34b-top50        & top50 type-checked random          & test  & 84.2       & 19.6     & 5.4   & 11.4  & 2.2       \\
    llemma-34b-top50        & top50 type-checked self consistent & test  & 84.2       & 28.3     & 9.8   & 14.7  & 2.2       \\
    llemma-34b-top50        & top20 type-checked self consistent & valid & 78.7       & 25.7     & 8.2   & 12.0  & 1.1       \\
    llemma-34b-top50        & top50 majority voting type-checked & valid & 24.6       & 10.9     & 4.9   & 7.1   & 0.0       \\
    llemma-34b-top50        & top50 random type-checked          & valid & 24.6       & 9.3      & 2.7   & 4.4   & 0.5       \\
    llemma-34b-top50        & top50 self consistent type-checked & valid & 32.8       & 14.2     & 3.8   & 6.0   & 0.0       \\
    llemma-34b-top50        & top50 type-checked majority voting & valid & 89.6       & 27.3     & 8.7   & 12.6  & 1.1       \\
    llemma-34b-top50        & top50 type-checked random          & valid & 89.6       & 21.3     & 4.9   & 9.8   & 2.2       \\
    llemma-34b-top50        & top50 type-checked self consistent & valid & 89.6       & 29.5     & 8.7   & 13.1  & 1.6       \\
    llemma-34b-top50        & top5 type-checked self consistent  & valid & 55.2       & 14.8     & 5.5   & 8.7   & 2.2       \\
    llemma-7b               & greedy type-checked                & test  & 29.9       & 10.9     & 5.4   & 6.5   & 2.2       \\
    llemma-7b               & greedy type-checked                & valid & 26.8       & 8.7      & 3.3   & 6.0   & 0.0       \\
    llemma-7b-mma           & valid                              & valid & 14.2       & 6.0      & 2.7   & 4.4   & 0.0       \\
    llemma-7b-mma-top50     & top50 type-checked most frequent   & valid & 61.2       & 10.9     & 3.8   & 5.5   & 1.6       \\
    llemma-7b-mma-top50     & top50 type-checked random          & valid & 61.2       & 9.3      & 4.4   & 5.5   & 2.7       \\
    llemma-7b-mma-top50     & top50 type-checked self consistent & valid & 61.2       & 13.7     & 4.9   & 6.6   & 2.7       \\
    llemma-7b-top50         & top50 type-checked majority voting & test  & 88.6       & 23.9     & 10.3  & 12.0  & 2.7       \\
    llemma-7b-top50         & top50 type-checked random          & test  & 88.6       & 21.2     & 9.8   & 11.4  & 4.9       \\
    llemma-7b-top50         & top50 type-checked self consistent & test  & 88.6       & 29.3     & 11.4  & 17.9  & 4.3       \\
    llemma-7b-top50         & top20 type-checked self consistent & valid & 71.0       & 19.7     & 7.1   & 10.4  & 1.6       \\
    llemma-7b-top50         & top50 majority voting type-checked & valid & 25.1       & 10.9     & 4.4   & 4.9   & 0.5       \\
    llemma-7b-top50         & top50 random type-checked          & valid & 25.7       & 6.0      & 3.3   & 3.8   & 1.1       \\
    llemma-7b-top50         & top50 self consistent type-checked & valid & 32.2       & 14.2     & 6.6   & 9.3   & 0.5       \\
    llemma-7b-top50         & top50 type-checked most frequent   & valid & 84.7       & 23.0     & 8.2   & 9.8   & 2.2       \\
    llemma-7b-top50         & top50 type-checked random          & valid & 84.7       & 16.9     & 4.9   & 7.7   & 2.7       \\
    llemma-7b-top50         & top50 type-checked self consistent & valid & 84.7       & 23.5     & 7.7   & 11.5  & 1.6       \\
    llemma-7b-top50         & top5 type-checked self consistent  & valid & 50.3       & 10.9     & 3.3   & 6.0   & 1.6       \\
    \bottomrule
\end{longtable}

\end{document}